%% file: main-arxiv.tex
\documentclass{article}

\usepackage[preprint]{venues/colm2026/colm2026_conference}

\usepackage[T1]{fontenc}
\usepackage{microtype}
\usepackage{graphicx}
\usepackage{subcaption}
\usepackage{booktabs}
\usepackage{multirow}
\usepackage{tabularx}
\usepackage{array}
\usepackage{hyperref}

\definecolor{darkblue}{rgb}{0, 0, 0.5}
\hypersetup{colorlinks=true, citecolor=darkblue, linkcolor=darkblue, urlcolor=darkblue}

\usepackage{colortbl}
\usepackage{amsmath}
\usepackage{amssymb}
\usepackage{mathtools}
\usepackage{amsthm}
\usepackage{enumitem}
\usepackage{makecell}
\usepackage{tikz}
\usepackage{pgfplots}
\pgfplotsset{compat=1.18}

\usepackage{xcolor}
\usepackage{fontawesome}
\newcommand{\fs}[1]{}  

\usepackage[capitalize,noabbrev]{cleveref}

\theoremstyle{plain}

\theoremstyle{definition}

\theoremstyle{remark}

\usepackage[disable]{todonotes}  
\usepackage{listings}
\makeatletter
\def\@maketitle{\vbox{\hsize\textwidth
\lhead{Preprint. Under review.}
\begin{center}
{\Large\bf \@title\par}
\vskip 0.4em
{\@author\par}
\end{center}
\vskip 0.3in minus 0.1in}}
\makeatother

\title{Can Current Agents Close the Discovery-to-Application Gap? A Case Study in Minecraft}

\author{
\textbf{Zhou Ziheng}$^{1*}$\quad
\textbf{Huacong Tang}$^{1*}$\quad
\textbf{Jinyuan Zhang}$^{1}$\quad
\textbf{Haowei Lin}$^{2}$\quad
\textbf{Bangcheng Yang}$^{1}$\\
\textbf{Qian Long}$^{3\dagger}$\quad
\textbf{Fang Sun}$^{1}$\quad
\textbf{Yizhou Sun}$^{1}$\quad
\textbf{Yitao Liang}$^{2}$\quad
\textbf{Ying Nian Wu}$^{1}$\\
\textbf{Demetri Terzopoulos}$^{1}$\quad
\textbf{Xiaofeng Gao}$^{3\dagger}$
\\[0.4em]
$^{1}$\textit{University of California, Los Angeles} \quad
$^{2}$\textit{Peking University} \quad
$^{3}$\textit{Amazon}
\\[0.3em]
$^{*}$Equal contribution.\quad
\texttt{josephziheng@ucla.edu}, \texttt{hctang@ucla.edu}
\\[0.6em]
\href{https://scicrafter-bench.github.io/}{\faGlobe~Project Page} \quad
\href{https://github.com/scicrafter-bench/scicraft-bench}{\faGithub~Code}
}\date{}

\newcommand{\scicrafter}{\textsc{SciCrafter}}

\begin{document}

\maketitle
\let\thefootnote\relax\footnotetext{$^{\dagger}$This work does not relate to the authors' position at Amazon.}

\begin{abstract}
\input{sections/0_abstract}
\end{abstract}

\input{figures/teaser_figure}
\input{sections/1_intro}

\input{sections/2_related_work}

\input{sections/3_benchmark}

\input{sections/4_methods}

\input{sections/5_experiments}

\input{sections/6_discussion}
\bibliography{bib/references}
\bibliographystyle{venues/colm2026/colm2026_conference}

\newpage
\appendix
\input{appendices/app_formal_definitions}
\input{appendices/app_limitations}
\input{appendices/app_tasks}

\input{appendices/app_exp_details}
\input{appendices/app_more_exp_results}
\input{appendices/app_env}
\input{appendices/app_prompt_templates}
\input{appendices/app_knowledge_book}
\input{appendices/app_failure_taxonomy}

\end{document}

%% file: sections/0_abstract.tex

Discovering causal regularities and applying them to build functional systems—the discovery-to-application loop—is a hallmark of general intelligence, yet evaluating this capacity has been hindered by the vast complexity gap between scientific discovery and real-world engineering. We introduce \scicrafter{}, a Minecraft-based benchmark that operationalizes this loop through parameterized redstone circuit tasks. Agents must ignite lamps in specified patterns (e.g., simultaneously or in timed sequences); scaling target parameters substantially increases construction complexity and required knowledge, forcing genuine discovery rather than reliance on memorized solutions. Evaluating frontier models including GPT-5.2, Gemini-3-Pro, and Claude-Opus-4.5 under a general-purpose code agent scaffold, we find that all plateau at approximately 26\% success rate. To diagnose these failures, we decompose the loop into four capacities—knowledge gap identification, experimental discovery, knowledge consolidation, and knowledge application—and design targeted interventions whose marginal contributions serve as proxies for corresponding gaps. Our analysis reveals that although the general knowledge application capability still remains as the biggest gap across all models, for frontier models the knowledge gap identification starts to become a major hurdle—indicating the bottleneck is shifting from \textit{solving problems right} to \textit{raising the right problems} for current AI. We release \scicrafter{} as a diagnostic probe for future research on AI systems that navigate the full discovery-to-application loop.

%% file: figures/teaser_figure.tex
\begin{figure*}[t]
    \centering
    \includegraphics[width=\linewidth]{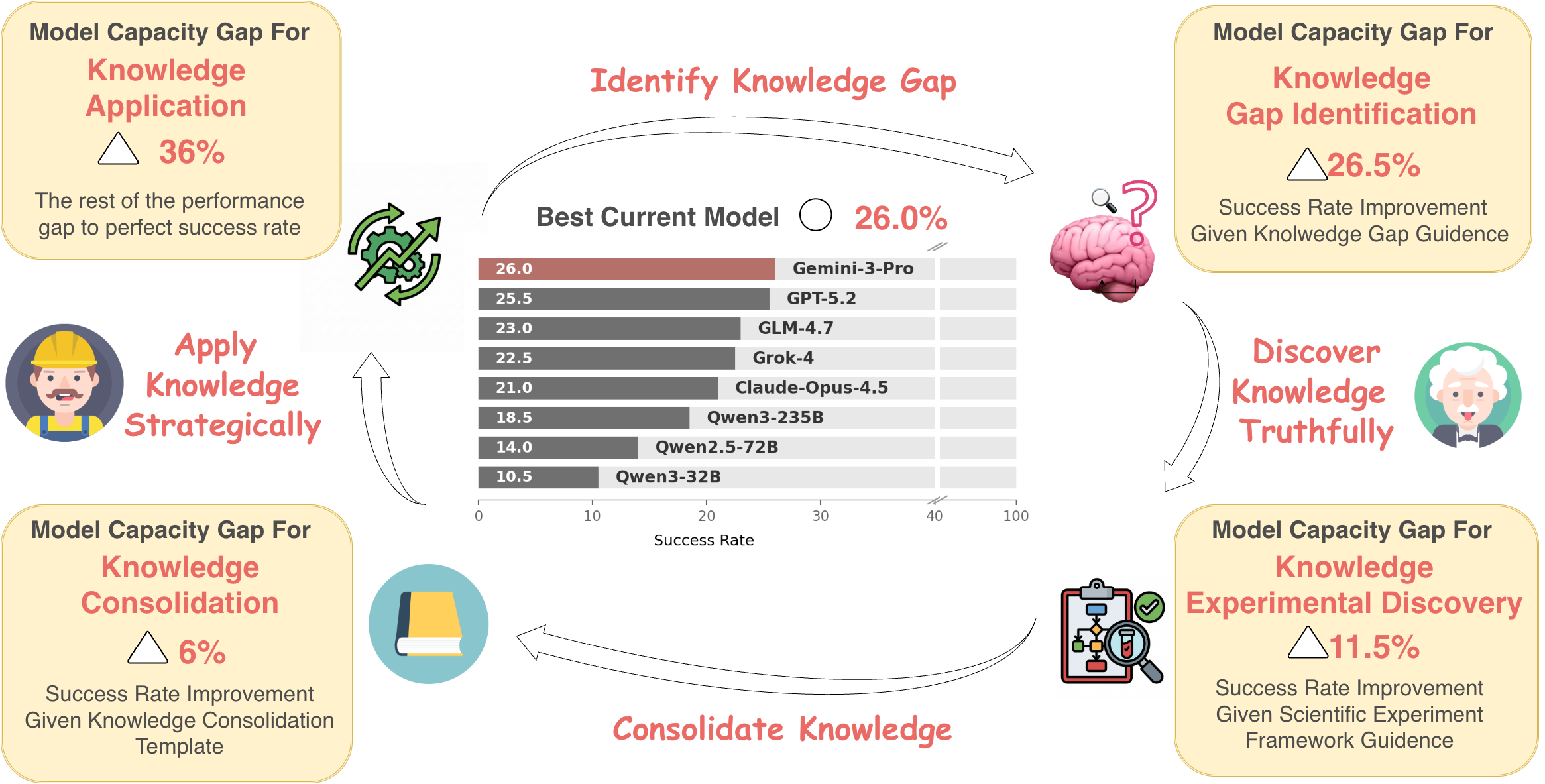}
    \caption{\textbf{Decomposing performance gaps in the Discovery-to-Application loop within \scicrafter{} (Gemini-3-Pro).} The best model achieves only 26.0\% success. We decompose the loop into four capacity gaps: \textbf{Knowledge Identification} (oracle hints on \textit{what} to discover boost success to 52.5\%), \textbf{Experimental Discovery} (a scientist sub-agent further reaches 64.0\%), \textbf{Knowledge Consolidation} (structured templates outperform free-form summaries), and \textbf{Application Capacity} (the remaining 36\% gap). See Table~\ref{tab:main_result_gap} for all models.}

    \vspace{-8pt}
    \label{fig:discovery_application_loop}
\end{figure*}

%% file: sections/1_intro.tex
\section{Introduction}
\label{sec:intro}



In human society, the interplay between discovery (e.g., science) and application (e.g., engineering) forms a self-evolving loop that drives technological advancement. Scientific investigations uncover causal models of how the world works~\citep{pearl2009causality,newell1976computer,gupta2024causalworldmodel}, providing knowledge for engineering to build complex systems~\citep{simon1996sciences}, while engineering challenges in turn spark new scientific inquiries. This capacity to navigate the discovery-to-application loop is a hallmark of general intelligence~\citep{lake2017building,tenenbaum2011how,gopnik2012reconstructing,goodman2008rational}. Indeed, anthropological evidence even suggests these intertwined processes played a pivotal role in the evolution of human cognition itself~\citep{Stout2011StoneToolmaking,Malafouris2021HowDoesThinkingRelate,Vaesen2012CognitiveBasesToolUse,LombardHogberg2021FourFieldCoevolution}.

A key barrier to evaluating the discovery-to-application loop is the ``complexity gap''—the vast disparity in timescale and resources between identifying a scientific principle and engineering a real-world application. Modern science-to-technology pipelines span years, and physical experiments require sophisticated robotic manipulation that remains unsolved.

Minecraft offers an ideal surrogate that bridges this gap. As a persistent 3D world with rich spatial, causal, and temporal dynamics, it supports open-ended construction of highly complex systems—up to fully functional CPUs.\footnote{\url{https://minecraft.fandom.com/wiki/Tutorials/Redstone_computers}} Crucially, while basic game mechanics are widely known, precise low-level details—such as signal interference patterns and structural constraints that vary across game editions—are difficult for LLMs to capture accurately. Agents therefore cannot rely solely on prior knowledge; they must discover these hidden dynamics to build functional devices. By abstracting physical manipulation into discrete block-based interactions, Minecraft isolates the core cognitive processes of scientific inquiry and engineering design from robotic limitations. Furthermore, the environment's rules can be programmatically modified, preventing agents from simply memorizing solutions and forcing genuine engagement with the discovery loop.


This setting allows us to investigate two fundamental questions: \textbf{(1) Can current AI agents autonomously navigate the full discovery-to-application loop---and if not, where is the gap? (2) Which cognitive capacities constitute the primary bottlenecks?} To address these questions, we introduce \scicrafter, a task suite that operationalizes and evaluates these capacities with scalable complexity.

\input{figures/task_illustration}

The task schema is intentionally simple: ignite $N$ lamps in specified patterns (e.g., simultaneously, or following a delay sequence $[t_1, t_2, \ldots, t_n]$) within a fixed area (see Figure~\ref{fig:task_illustration}). This design ensures evaluation remains invariant across difficulty levels---complexity scales by adjusting target parameters alone---while construction complexity and required knowledge grow substantially. For instance, synchronizing four lamps may succeed with basic wiring, but scaling to eight requires discovering the ``nested hub'' pattern. Scaling further from 32 to 64 lamps demands understanding signal degradation and how repeaters circumvent it.

We design five task variations: (1) Simultaneous Ignition, (2) Branch Reach (T-Junction Expansion), (3) Sequential $N$-Stage Activation, (4) Equal-Delay Distribution, and (5) Pulse Extension. Each variation includes five manually calibrated difficulty levels with balanced knowledge gaps, though users can configure arbitrary difficulty settings.


To assess how modern AI navigates this discovery-to-application loop, we evaluated a suite of state-of-the-art models, ranging from frontier reasoning models such as
GPT-5.2, Gemini-3-Pro, and Claude-Opus-4.5 to popular small open-source models such as Qwen3-32B. All models were equipped with a general-purpose coding agent scaffold
(Claude Code) to provide a standardized harness. We chose this setup for three reasons: bare-bone LLMs lack the tool use and memory capabilities required for a complex
task like ours; manually engineered scaffolds (custom memory structures, workflows, etc.) introduce confounding variables that make it difficult to draw stable
conclusions; and coding agent scaffolds are widely adopted for general agentic tasks, making them a reasonable proxy for current LLM capability — though not an upper
bound on what current AI can achieve. Under this setup, all agents plateau at approximately 26\% success rate (\cref{fig:discovery_application_loop}).


To diagnose the performance bottlenecks responsible for this failure, we decompose the discovery-to-application loop into four distinct capacities: (1) \textbf{Knowledge Gap Identification}---the ability to identify knowledge gaps and formulate targeted research questions; (2) \textbf{Experimental Discovery}---the capacity to design and execute rigorous experiments to infer unobservable causal mechanisms; (3) \textbf{Knowledge Consolidation}---the ability to distill findings into concise, reusable forms for future application; and (4) \textbf{Knowledge Application}---the foundational ability to reason, plan, and execute precise engineering, defined as the residual capacity not covered by the above.

To isolate each capacity, we devise a series of ``oracle'' interventions whose marginal contributions serve as proxies for the corresponding capacity gaps. For knowledge gap identification, we provide manually crafted hints indicating areas of missing knowledge (e.g., ``signal flow direction'') without revealing specifics---models know \textit{what} to discover but not the answer. For experimental discovery, we introduce a ``scientist'' sub-agent to guide more rigorous investigation. For knowledge consolidation, we design a structured knowledge entry format. The design details are in Section~\ref{sec:agent-framework}. Note that these agent and consolidation designs are simultaneously contributions of agent methods. The remaining gap is attributed to knowledge application by definition.  

Our analysis reveals that the general knowledge application is still currently the major gap for all models, but for frontier models the knowledge gap identification start to gain dominance. Also our ``scientist'' sub-agent and consolidation methods yield 0.33--1.00$\times$ relative gains, revealing significant untapped potential in experimental discovery.

In summary, our contributions are three-fold:
\begin{enumerate}[nosep, leftmargin=*]
    \item We construct \scicrafter{}, a set of discovery-to-application-loop tasks in Minecraft with automatically scalable difficulty.
    \item We decompose the capacity gap of LLMs in this loop into four components and evaluate a wide range of current language models.
    \item We contribute a ``scientist'' sub-agent and a knowledge consolidation method that significantly improve an agent's discovery ability.
\end{enumerate}

%% file: figures/task_illustration.tex
\begin{figure*}[!t]
    \centering
    \includegraphics[width=\linewidth]{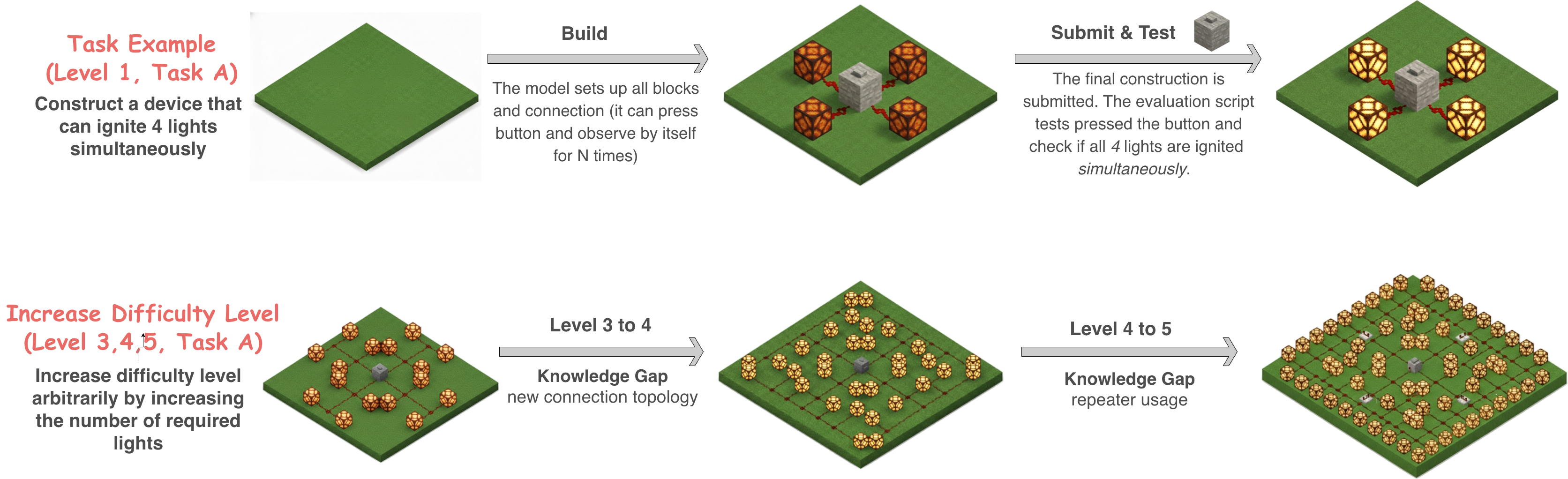}
    \caption{\textbf{\scicrafter Task Design Illustration.} 
    \textbf{Top (Task Procedure):} The model is tasked with constructing a functional device within a constrained vacant area based on provided instructions. During construction, the agent can interact with the device (e.g., by pressing a button) and observe its behavior to iterate on the design. Finally, the device is evaluated by an automated script that verifies if the output lighting patterns match the target specifications.
    \textbf{Bottom (Task Difficulty):} Task complexity is parametrically scalable by adjusting the required number of lights ($N$). For some temporal tasks, difficulty is further increased by requiring specific sequential patterns defined by a parameter array $[N_1, N_2, \dots, N_M]$. See details in the Appendix \ref{app:tasks}.}
    \label{fig:task_illustration}
    \vspace{-8pt}
\end{figure*}

%% file: sections/2_related_work.tex
\section{Related Works}

\paragraph{Language model agents.}
Recent work has turned LLMs into \emph{tool-using} agents by interleaving natural-language reasoning with explicit actions or API calls, including prompting-time agent loops and training-time tool-use objectives \cite{yao2022react,schick2023toolformer,karpas2022mrkl,qin2023toolllm}. A parallel line studies \emph{long-horizon planning} and hierarchical control with LLM-generated decompositions or search over intermediate thoughts/plans \cite{yao2023treeofthoughts,ahn2022saycan,huang2022zeroshotplanners}. For \emph{memory, reflection, and self-improvement}, agents maintain external state and iteratively critique/refine their own outputs, enabling multi-session competence beyond a single context window \cite{shinn2023reflexion,madaan2023selfrefine,packer2023memgpt,park2023generativeagents}. Finally, multi-agent and role-specialization patterns (e.g., scientist/engineer sub-agents that converse and delegate) provide a practical abstraction for complex workflows \cite{wu2023autogen,li2023camel,hong2023metagpt}, and agent evaluations increasingly target interactive, multi-step settings and real software tasks \cite{liu2023agentbench,jimenez2024sweagent}.

\paragraph{Automated scientific discovery and experimentation.}
Beyond assisting scientists, recent systems attempt end-to-end automation of the research loop---idea generation, implementation, experimentation, and paper writing---as exemplified by fully autonomous ``AI scientist'' pipelines \cite{lu2024aiscientist,song2025evaluating,jansen2024discoveryworld}. In chemistry, LLM-driven autonomous research agents integrate tool use, planning, and execution for closed-loop discovery \cite{boiko2023autonomous}. Complementary work targets discovery in algorithmic domains by coupling LLMs with search or evolutionary procedures to yield novel results \cite{romeraparedes2024funsearch}. These efforts connect naturally to \emph{autonomous experimentation} and \emph{active experimental design} in self-driving labs, where Bayesian optimization and active learning select informative experiments under cost and noise constraints \cite{hase2019selfdrivinglabs,hase2018phoenics,shahriari2016bayesopt,settles2009activelearning}. For scientific settings where \emph{causal structure} matters, interventional causal discovery provides principled objectives and guarantees, ranging from classical frameworks to modern scalable optimization-based structure learning \cite{pearl2009causality,spirtes2000cps,hauser2012gies,peters2016icp,zheng2018notears}.

\paragraph{Minecraft as testbed.}
Minecraft has emerged as a rich, controllable sandbox for embodied intelligence, supporting both reinforcement learning and interactive agent evaluation. Foundational platforms and datasets enable reproducible experimentation at scale \cite{johnson2016malmo,guss2019minerl}, while newer frameworks broaden task diversity and incorporate internet-scale knowledge for open-ended goals \cite{fan2022minedojo,wang2023voyager}. Recent approaches learn general behavioral priors from unlabeled human gameplay video \cite{baker2022vpt} and adapt them to instruction-following with text-conditioned behavior generation \cite{lifshitz2023steve1}; dialogue-grounded building assistants further support interactive construction tasks \cite{gray2019craftassist,qian2024teamcraft}. Benchmarking suites for open-ended game agents aim to systematically measure generalization across heterogeneous Minecraft tasks \cite{zheng2025mcu}. While Minecraft contains complex circuit-like mechanics (redstone), explicit benchmarks for redstone \emph{reasoning} remain limited; existing work more often treats redstone as a stylistic/structural building component rather than as a dedicated logical reasoning target \cite{huang2023redstonecities}.

%% file: sections/3_benchmark.tex
\section{Task Construction}
\paragraph{High Level Task Schema}
We design \scicrafter{} with two principles in mind: \emph{systematic curriculum} in knowledge gaps and \emph{automated scalability}. Tasks must pose a structured sequence of knowledge gaps that require genuine discovery, and task complexity must scale automatically without manual redesign of each difficulty level.

To satisfy these desiderata, we adopt a simple yet expressive task schema: ignite $N$ lamps in specified patterns (e.g., simultaneously, or following a delay sequence $[t_1, t_2, \ldots, t_n]$) within a fixed area. This design ensures that evaluation remains invariant across difficulty levels, while the knowledge gap to meet the requirements consistently grow, necessitating the discovery of new environmental mechanics. For instance, an agent may successfully synchronize four lamps with basic wiring, but scaling to eight requires discovering the ``nested hub'' pattern; scaling further from 32 to 64 lamps demands understanding signal degradation and how repeaters can circumvent it (see Figure~\ref{fig:task_illustration}).

\paragraph{Task Families and Knowledge Gaps}

We design five task families that probe distinct spatial and temporal constraints:
\textbf{(A) Simultaneous Ignition}---activate $N$ lamps at the same tick;
\textbf{(B) T-Junction Routing}---connect lamps using a trunk-and-branch layout;
\textbf{(C) Sequential Activation}---activate lamps with specified inter-stage delays $[t_1, t_2, \ldots, t_n]$;
\textbf{(D) Distance-Equalized Ignition}---simultaneously activate lamps placed at heterogeneous distances; and
\textbf{(E) Pulse Extension}---maintain activation for a specified duration $\tau$.
Details are provided in Appendix~\ref{app:tasks}.

Crucially, difficulty does not increase smoothly with task parameters (e.g., $N$ in Family~A); instead, higher levels cross discrete \emph{mechanism thresholds} that require discovering new redstone principles. We identify three core knowledge dimensions. The first is \textbf{local wiring grammar}: dust propagates only through axis-aligned adjacency (no diagonal conduction), auto-connects to neighbors (creating unintended branches), and must physically contact a lamp to power it; a button strongly powers its supporting block, so omitting the block yields a plausible but non-functional circuit. The second is \textbf{attenuation-aware topology}: dust carries strength $\in\{0,\ldots,15\}$ that decays by one per block and vanishes after 15 segments, forcing hub/backbone designs or explicit regeneration for larger footprints. The third is \textbf{repeater semantics}: repeaters regenerate signal to full strength but act as directional diodes with 1--4 ticks latency; this latency accumulates over chains, making timing skew a first-class constraint, and side power can lock repeaters unexpectedly.

Each family surfaces different knowledge gaps. For Family~A : early levels admit symmetric branching, mid levels require denser topologies within the attenuation budget, and high levels force repeater insertion with joint delay-distance balancing (see Figure~\ref{fig:task_illustration}). Family~B enforces T-junction layouts, turning the problem into topology-constrained routing. Family~C requires composing quantized repeater delays into precise delay lines. Family~D demands using repeaters as compensatory delay elements to equalize heterogeneous path lengths. Family~E adds duration constraints requiring pulse-shaping techniques.

Each task is defined by a tuple $m = (I, s_0, u, \varphi)$: a natural-language instruction $I$, an initial world state $s_0$, a stimulus action $u$ (e.g., pressing a button), and a temporal checker $\varphi$ that verifies whether the built artifact produces the desired output pattern. The agent constructs a device, and the evaluator applies the stimulus and records the resulting per-tick state trace, declaring success iff $\varphi(\mathbf{s})=1$. In practice, $I$ is a YAML task description, while $u$ and $\varphi$ are executable scripts. The formal mathematical definitions are provided in Appendix~\ref{app:formal_definitions}.

\paragraph{Environment-Agent Interface (MCP)}
To facilitate the community to test our tasks, we wrap the interaction interface between the environment and the agent using  \textit{Model Context Protocol (MCP)}\footnote{\url{https://modelcontextprotocol.io}}. MCP has become the standard function call protocol for LLMs so the capability of LLMs would not be affected by the function call format.  Please see Appendix~\ref{app:env} for specific MCP calls.

%% file: sections/4_methods.tex
\section{\textit{Scientist} Sub-agent And Knowledge Consolidation}
\label{sec:agent-framework}

As mentioned in the introduction, we provided a ``scientist'' sub-agent method to both facilitate the evaluation and as a general method contribution. The sub-agent serves as a way for the main agent to discover new knowledge through experimentation. Every time the main agent believes it encounters a knowledge gap to fill, it simply prompts the sub-agent with a question to investigate---for
instance, \emph{``How long does a stone button remain pressed after 
activation?''} And then the scientist sub-agent will not rely on prior assumptions, but instead conduct systematic control experiments to find out. When it finishes, it updates the knowledge book shared with the main agent. The main agent can then reattempt the construction task using the refined knowledge, and prompt new questions to the sub-agent.

\subsection{\textit{Scientist} Sub-agent Design}
\label{subsec:scientist}

The major design for the scientist agent is providing it with a template for conducting experiments and a template for extracting knowledge as the system prompt (full prompts and templates are in Appendix~\ref{app:prompt_templates}). Along with them, it is also provided with the current knowledge book, the history of past experiment reports, and a basic wiki of Minecraft from online. 
  
This scientific experiment template design is drawn upon AI Scientists line of works \citep{lu2025aiscientistv2, jansen2024discoveryworld, elahi2024adaptive}. The essence is to formulate hypothesis, design experiments, analyze results and iterate. We break them down to eight specific parts: (1)~\textit{Research Question}---identify the specific mechanic under investigation; (2)~\textit{Hypothesis}---formulate a testable prediction; (3)~\textit{Experiment Design}---specify the independent variable, measurement target, and controls; (4)~\textit{Experiment Steps}---describe procedures to execute each trial; (5)~\textit{Experiment Record}---document observations and note hypothesis alignment; (6)~\textit{Experiment Results}---summarize empirical outcomes; (7)~\textit{Analysis \& Summary}---interpret patterns and evaluate the hypothesis; and (8)~\textit{Next Steps}---propose follow-up experiments to refine the discovered law. This workflow is provided as a system prompt replacement of the base code agent rather than enforced as a rigid execution script. The system leverages the LLM and code agent's general capability to follow the structure while allowing flexibility in how each step is executed. Note that we asked the scientist agent to repeat each experiment three times to ensure reliable results.

This eight-part structure also serves as the template for the written experiment report that the scientist agent must complete after each experimental iteration.  
Upon finishing one set of experiments, an experiment report is created and archived for future reference. Then, based on this new experiment report, combining the complete experiment history and the existing entries in the knowledge book, the scientist agent updates the knowledge book with the new finding.

\subsection{Knowledge Consolidation Structure}
\label{subsec:knowledge_book}
After the experiments, the scientist agent consolidates its findings into the knowledge book. We find this structure is critical for later performance (see Table~\ref{tab:knowledge_consolidation_ablation}). The final design is a four-part structure comprising:
\begin{enumerate}[nosep, leftmargin=*]
    \item \textit{Claim (law):} The discovered law or dynamics statement.
    \item \textit{Evidence Proof:} The proof based on experiments.
    \item \textit{Constraints:} The constraints of application for the found law.
    \item \textit{Example:} A practical example of how to apply it.
\end{enumerate}
A claim in the knowledge book is like ``Redstone signal strength decreases by 1 for every block of distance traveled''. To see the prompt, please see Appendix~\ref{sec:appendix-knowledge-book} for details. Note that we find that the knowledge book structure affects the final performance non-trivially. Please see the ablation experiment in later section. 

At each discovery iteration $k$, the scientist agent takes a query $Q_k$, conducts an experiment trajectory, generates an experiment report $ER_k$, and updates the knowledge book $KB_k$ incrementally. This ensures that the scientist's evolving understanding is reflected in real-time as each iteration concludes. The formal mapping is provided in Appendix~\ref{app:formal_definitions}.

%% file: sections/5_experiments.tex
\section{Experiments}
\vspace{-5pt}
\label{sec:Experiments}

\subsection{Experiment Design Method} 
\label{sec:experiment_design}
To diagnose the capability gap within the discovery-to-application loop, we decompose the procedure into four steps grounded in common science and engineering practice:
(1) knowledge gap identification — recognizing what needs to be discovered; (2) knowledge discovery through experiments — conducting systematic experiments to collect
evidence, verify hypotheses, and refine understanding; (3) knowledge consolidation — presenting and preserving findings in a reusable form; and (4) the residual, which
we term knowledge application — the general capacity to understand, reason about, and apply existing knowledge. We acknowledge that this is not the only valid
decomposition. One could alternatively analyze capability gaps through the lens of spatiotemporal reasoning, long-context management, or other constituent abilities.
Here, we focus on a decomposition aligned with the discovery-to-application procedure itself.

To quantify these gaps, we use the \textit{marginal contribution} of targeted scaffolding interventions as proxies. Because each intervention simultaneously provides assistance and alters agent behavior (e.g., hints may also serve as metacognitive cues signaling that exploration is needed), the measured gaps reflect the joint effect of the intervention and the model's response to it, rather than cleanly isolating a single latent capacity.

For the knowledge identification stage, we provide high-level oracle hints (e.g., `signal flow direction') that guide the model toward the relevant area of investigation without specifying the underlying mechanisms. This allows us to separate the model's ability to identify a gap from its ability to discover it.

The \textit{scientist} sub-agent and knowledge consolidation methods are detailed in Sections~\ref{subsec:scientist} and~\ref{subsec:knowledge_book}. Notably, the sub-agent requires a consolidation mechanism to relay discoveries to the primary agent, precluding fully isolated evaluation of discovery capacity. Since consolidation filters and organizes information rather than generating it, the consolidation method acts as an upper bound on the sub-agent's efficacy. We therefore pair the sub-agent with our optimized consolidation method when measuring the discovery capacity gap, providing a more accurate estimate of each model's intrinsic discovery potential.

We formalize the performance of a model $M$ as the success probability $P(S{=}1 \mid M, \mathcal{A})$, where $\mathcal{A}$ is a set of contextual assistances. We define four gaps as the marginal gains from sequentially introducing each intervention: (1)~\textbf{Knowledge Identification Gap} ($\delta_{id}$): the gain from providing oracle hints over the unassisted baseline; (2)~\textbf{Knowledge Discovery Gap} ($\delta_{ds}$): the additional gain from introducing the scientist sub-agent with optimized consolidation; (3)~\textbf{Consolidation Optimization Gap} ($\delta_{kc}$): the gain from switching to a structured consolidation template; and (4)~\textbf{Application Gap} ($\delta_{app}$): the residual gap to perfect performance, representing foundational capabilities such as spatial reasoning and code generation that our interventions do not directly address. These four gaps, together with the baseline success rate, partition the total capacity space to 100\%. We also report the \textit{relative gap ratio} $r_\delta = \delta / P(S{=}1 \mid M, \emptyset)$ to normalize across models with different baselines. Formal definitions are in Appendix~\ref{app:formal_definitions}.

\subsection{Experiment Setup}

We evaluate a diverse suite of LLMs---GPT-5.2, Claude-Opus-4.5, Gemini-3-Pro, Grok-4, GLM-4.7, Qwen3-235B-MoE, Qwen2.5-72B-Instruct, and Qwen3-32B---using Claude Code~\citep{anthropic2025claudecode}, a state-of-the-art code agent framework. Each model is given a budget of 50 verification trials per task and evaluated in a curriculum setting (L1$\to$L5) with experience carried forward. The definition of one verification trial is one time that the agent presses the button and receive environmental feedback. Success rate is averaged over eight runs. Full details are provided in Appendix~\ref{app:exp_details}.

\section{Results and Discussion}
\label{sec:results}

\subsection{Baseline Performance}

Table~\ref{tab:main_result_gap} summarizes results across all models. The best model, Gemini-3-Pro, achieves only 26.0\% success, with all frontier models plateauing near this level despite parameter counts ranging from 72B to an estimated 1.7 trillion for Grok-4. This suggests that scaling model size alone does not resolve the fundamental bottlenecks in the discovery-to-application loop.

\definecolor{lightgray}{gray}{0.9}

\begin{table*}[t]
    \centering
    \footnotesize
    \setlength{\tabcolsep}{2pt}
    \renewcommand{\arraystretch}{1.2}

    \resizebox{\textwidth}{!}{
    \begin{tabular}{l c >{\columncolor{lightgray}}c c >{\columncolor{lightgray}}c c >{\columncolor{lightgray}}c}
    \toprule
        \textbf{Model} & \textbf{Baseline} &
        \textbf{\shortstack{Know. Iden. Gap \\ ($\delta_{id}, r_{\delta_{id}}$)}} &
        \textbf{w/ Hint} &
        \textbf{\shortstack{Discovery Gap \\ ($\delta_{ds}, r_{\delta_{ds}}$)}} &
        \textbf{w/ Hint + Scientist} &
        \textbf{\shortstack{Residual (App Gap) \\ ($\delta_{app}, r_{\delta_{app}}$)}} \\
        \midrule
        \texttt{gemini-3-pro}         & $26.0$ & $\Delta 26.5$ ($1.02\times$) & $52.5$ & $\Delta 11.5$ ($0.44\times$) & $64.0$ & $\Delta 36.0$ ($1.38\times$) \\
        \texttt{gpt-5.2}               & $25.5$ & $\Delta 25.5$ ($1.00\times$) & $51.0$ & $\Delta 9.0$ ($0.35\times$) & $60.0$ & $\Delta 40.0$ ($1.57\times$) \\
        \texttt{claude-opus-4.5}       & $21.0$ & $\Delta 25.0$ ($1.19\times$) & $46.0$ & $\Delta 13.0$ ($0.62\times$) & $59.0$ & $\Delta 41.0$ ($1.95\times$) \\
        \texttt{glm-4.7}               & $23.0$ & $\Delta 22.5$ ($0.98\times$) & $45.5$ & $\Delta 7.5$ ($0.33\times$) & $53.0$ & $\Delta 47.0$ ($2.04\times$) \\
        \texttt{grok-4}                & $22.5$ & $\Delta 20.0$ ($0.89\times$) & $42.5$ & $\Delta 14.0$ ($0.62\times$) & $56.5$ & $\Delta 43.5$ ($1.93\times$) \\
        \texttt{qwen3-235b}            & $18.5$ & $\Delta 24.0$ ($1.30\times$) & $42.5$ & $\Delta 13.0$ ($0.70\times$) & $55.5$ & $\Delta 44.5$ ($2.41\times$) \\
        \texttt{qwen2.5-72b}           & $14.0$ & $\Delta 15.0$ ($1.07\times$) & $29.0$ & $\Delta 14.0$ ($1.00\times$) & $43.0$ & $\Delta 57.0$ ($4.07\times$) \\
        \texttt{qwen3-32b}             & $10.5$ & $\Delta 27.0$ ($2.57\times$) & $37.5$ & $\Delta 9.0$ ($0.86\times$) & $46.5$ & $\Delta 53.5$ ($5.10\times$) \\
        \bottomrule
    \end{tabular}
    }

    \caption{\textbf{Model Performance and Gap Decomposition (Curriculum Setting).} Success rates (\%) aggregated across all 25 tasks (5 families $\times$ 5 levels) over 8 independent runs (i.e., $k/200 \times 100$). Gray columns show the marginal performance gain ($\delta$) from each intervention and its ratio to the baseline ($r_{\delta}$). Absolute gaps ($\delta$) decompose the 100\% performance space; ratios ($r_{\delta}$) normalize for baseline differences but are subject to ceiling/floor effects.}
    \label{tab:main_result_gap}
\end{table*}

\input{figures/fig_failure_cases}

\subsection{Diagnosing Performance Gaps}

We conducted systematic ablations using targeted scaffolding interventions (Table~\ref{tab:main_result_gap}). As noted in Section~\ref{sec:experiment_design}, these interventions serve as proxies; the measured gaps reflect the marginal benefit of each form of assistance.

\textbf{Knowledge Gap Identification.} Oracle hints (e.g., ``signal decay ratio'') that guide discovery without revealing solutions yield the largest single improvement: all models roughly double their success rate, with absolute gains of 15.0--27.0\%.

\textbf{Experimental Discovery.} Adding a ``scientist'' sub-agent with optimized knowledge consolidation provides further gains of 7.5--14.0\% absolute across all tiers, bringing the best configuration (Gemini-3-Pro) to 64.0\%.

\textbf{Knowledge Consolidation.} The consolidation format proves critical (Table~\ref{tab:knowledge_consolidation_ablation}). Unstructured summarization captures less than half the gain achievable with proper consolidation. Our structured ``Claim-Proof-Constraints-Example'' format---recording findings as law-like claims with evidence, constraints, and examples---achieves 64.0\%, substantially outperforming free-form summaries at 58.0\% (see Appendix~\ref{sec:appendix-knowledge-book}).

\textbf{Knowledge Application.} The residual gap (36.0--57.0\%) encompasses spatial reasoning, code generation, and long-horizon context management---capabilities our interventions do not directly address. This gap scales inversely with model capability, ranging from 36.0\% for Gemini-3-Pro to 57.0\% for Qwen2.5-72B.

\textbf{Additional Results.} Curriculum learning (L1$\rightarrow$L5) outperforms independent evaluation (Appendix~\ref{app:more_exp_results}). Even with full support, no model succeeds at Level~5.

\paragraph{Qualitative Failure Analysis.} We observed and summarized 12 failure modes from different models' execution. These failure modes cluster into three categories that mirror our capacity decomposition: \emph{structural failures} (e.g., reversed repeaters blocking signal propagation) map to the knowledge application gap; \emph{signal propagation failures} (e.g., long serial paths without amplification) map to the discovery gap; and \emph{wire semantics failures} (e.g., directional connection mismatches) map to the identification gap---the most subtle class, producing circuits that \emph{appear correct} yet remain functionally broken. This progression from obvious to subtle errors mirrors the capacity gap hierarchy in our quantitative results. We display one device illustration in (Figure~\ref{fig:failure_example}). The complete taxonomy along with device snapshot illustration is in Appendix~\ref{app:failure_taxonomy}.
  
\subsection{Discussion}

Our results show that the general knowledge application capability gap is rapidly shrinking for frontier models. The knowledge identification gap start to be comparatively dominating. While it is initially surprising to see its important significance, the finding aligns with Einstein's observation that formulating a problem is often more essential than solving it~\cite{einstein1938evolution}---posing effective questions demands discerning which areas are most promising. The experimental discovery gap is also notable: the ``scientist'' agent follows a generalized procedure that should already reside within LLM prior knowledge, yet requiring models to follow a formal experimental structure yields 0.33--1.00$\times$ gains, suggesting autonomous discovery capabilities remain underdeveloped. Moreover, the stark differences among consolidation formats reveal that LLMs perform poorly at determining how to store knowledge. The ``Claim-Proof-Constraints-Example'' format outperforms the intuitive ``Finding-Explanation-Example'' format, likely because delineating conditions under which claims hold enables better understanding of when to apply them---offering insights for memory evolution research~\citep{hu2025memory,xu2025mem}. 

\subsection{Limitations and Future Work}

First, our results reflect the joint performance of the model and the coding agent scaffold. We believe this combination provides a reasonable representation of current general-purpose LLM agent capabilities, though it should not be regarded as an upper bound. Second, while the Minecraft environment offers convenience, efficiency, controllability, and scalable difficulty, it does not capture all the complexities of real-world discovery. Accordingly, our proposed tasks should be viewed as a diagnostic probe for assessing model capabilities across the complex loop of scientific discovery and application, rather than as a definitive benchmark. Third, our four-way decomposition employs interventions whose effects are not fully orthogonal; the measured gaps should therefore be interpreted as marginal contributions and quantitative diagnostic signals rather than isolated capacity measurements. In the future, We will incorporate vision input to assess multimodal capabilities. We also plan to support randomization of the underlying environment dynamics to prevent solutions based on memorization. A detailed discussion of these limitations and future directions is provided in Appendix~\ref{app:limitations}.

\section{Conclusion}
\label{sec:conclusion}

We introduce \scicrafter{}, a Minecraft-based benchmark evaluating language model agents in discovery-to-application scenarios where even the best models achieve only 26\% success. Using diagnostic interventions, we decompose performance gaps into knowledge gap identification, experimental discovery, knowledge consolidation, and knowledge application. While most models are primarily limited by application capacity, frontier models are increasingly bottlenecked by knowledge gap identification, indicating the main challenge starts to shift from \textit{solving problems right} to \textit{raising right problems}. We release \scicrafter{} as an open diagnostic testbed for evaluating diverse agent architectures.

%% file: figures/fig_failure_cases.tex
\begin{figure*}[t]
  \centering
  \begin{minipage}[t]{0.50\linewidth}
    \vspace{0pt}
    \centering
    \small
    \renewcommand{\arraystretch}{1.4}
    \begin{tabular*}{\linewidth}{@{\extracolsep{\fill}}l c@{}}
      \toprule
      \textbf{Knowledge Consolidation Structure} & \textbf{w/Hint +} \\
      & \textbf{Scientist} \\
      \midrule
      Self-determined Summary & $58.0$ \\
      \texttt{Finding-Explanation-Example} & $60.5$ \\
      \texttt{Claim-Proof-Constraints-Example} & $64.0$ \\
      \bottomrule
    \end{tabular*}
    \renewcommand{\arraystretch}{1.0}
    \captionof{table}{\textbf{Comparison of knowledge consolidation methods.} Success rates (\%) measured using Gemini-3-Pro with the scientist sub-agent and hints (8 runs). The results show that the choice of consolidation structure critically affects downstream performance.}
    \label{tab:knowledge_consolidation_ablation}
  \end{minipage}\hfill
  \begin{minipage}[t]{0.48\linewidth}
    \vspace{0pt}
    \centering
    \includegraphics[width=0.85\linewidth]{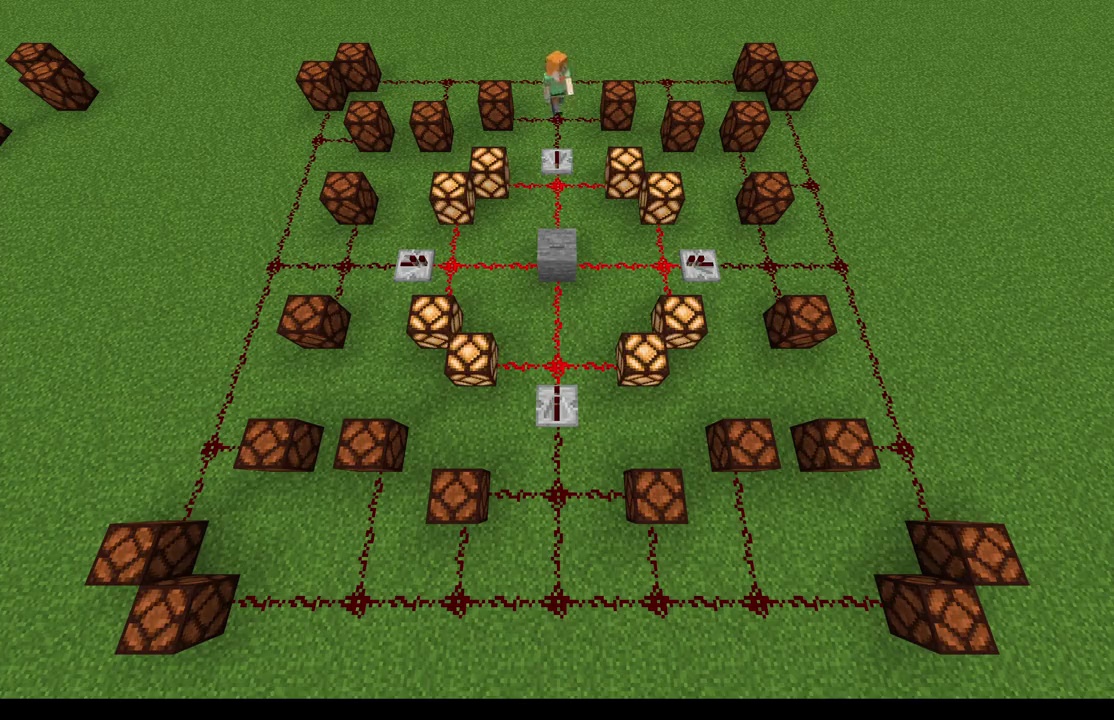}
    \vspace{-0.5em}
    \captionof{figure}{\textbf{A representative failure case} from the 32-lamp task. Repeaters oriented backwards block signal to 24 of 32 lamps. See Appendix~\ref{app:failure_taxonomy} for the full taxonomy.}
    \label{fig:failure_example}
  \end{minipage}
\end{figure*}

%% file: appendices/app_formal_definitions.tex
\section{Formal Definitions}
\label{app:formal_definitions}

\subsection{Task Formalization}

Let $\mathcal S$ denote the Minecraft world-state space. We write the per-tick
environment transition as $\tau(\cdot \mid s)$, i.e., $s_{t+1} \sim \tau(\cdot \mid s_t)$.
Our suite of tasks is denoted as $\mathcal M$. Each task is represented by a tuple
\[
m = (I, s_0,\; u,\; \varphi)\in\mathcal M,
\]
where $I$ is the task instruction provided to AI agent, $s_0\in\mathcal S$ is the initial world state, $u:\mathcal S\to\mathcal S$ is the stimulus for triggering the test (generally applied at evaluation start, e.g., pressing a button), and $\varphi$ is a temporal checker that checks the recorded per-tick test trace $\mathbf{s}$ and determines whether the artifact achieves desired functions and constraints under the stimulus.

Specifically, given $(I,s_0)$, the agent follows its build policy $\pi$ and submits a final world state $s^\star$, which is the terminal state of the induced rollout under the environment dynamics:
\[
(a_{0:T-1},\, s^\star) \sim \pi(\cdot \mid I, s_0),
\]
where $a_{0:T-1}$ denotes the sequence of build actions taken by the agent and $T$ is the (agent-determined) termination step.

Then given this submitted final world state $s^\star$, the evaluator generates a finite
test trace $\mathbf{s}=(s^{\text{test}}_0,s^{\text{test}}_1,\ldots)$ by
\[
s^{\text{test}}_0 = u(s^\star),\qquad
s^{\text{test}}_{t+1}\sim \tau(\cdot\mid s^{\text{test}}_t),
\]
and declares success iff $\varphi(\mathbf{s}) = 1$,
where any constraints (e.g., spatial distance, allowed components) and function tolerance threshold (e.g., ``simultaneous within $\le 1$ tick'') is encoded inside $\varphi$.

In practice, task instruction $I$ is a YAML-format text description provided to the LLM agent, and $u, \varphi$ are executable build-test scripts that trigger the built artifact and evaluate the recorded state changes to check function and constraint satisfaction.

\subsection{Scientist Agent Formulation}

We formulate the \textit{scientist agent} as a hierarchical mapping operating across two distinct temporal scales: the environment step $t$ and the discovery iteration $k$. For a given query $Q_k$, the agent conducts an experiment consisting of a trajectory $\mathcal{H}_{T}$ of $T$ environment steps. After the trajectory is complete, the agent immediately generates an experiment report $ER_k$ and updates the knowledge book $KB_k$. The process for each iteration $k$ is defined as:

\[
\text{F}_{\text{Scientist}} : (KB_{k-1}, Q_k, W, ER_{1:k-1}, \mathcal{H}_{T})
\longrightarrow (\{A_t\}_{t=1}^T, ER_k, KB_k)
\]

where $KB_k$ is the knowledge book updated incrementally based on the latest findings. Each $ER_k$ is an individual experiment report generated after a trajectory of experiment actions $\mathcal{H}_{T} = \{(A_t, O_t)\}_{t=1}^T$. This ensures that the scientist's evolving understanding is reflected in the knowledge book in real-time as each discovery iteration concludes.

\subsection{Capacity Gap Definitions}

We formalize the performance of a model $M$ as the success probability $P(S=1 \mid M, \mathcal{A})$, where $S \in \{0, 1\}$ is a binary random variable indicating task completion and $\mathcal{A}$ is a set of contextual assistances. The four capacity gaps are defined as follows:

\begin{enumerate}
    \item \textbf{Knowledge Identification Gap ($\delta_{id}$)}: Measured as the gain achieved by providing oracle identification guidance over the baseline:
    \begin{equation}
        \delta_{id} = P(S=1 \mid M, \{h_{id}\}) - P(S=1 \mid M, \emptyset)
    \end{equation}

    \item \textbf{Knowledge Discovery Gap ($\delta_{ds}$)}: The gain from further introducing the scientific sub-agent that specializes at doing scientific control experiments. Since it must use one consolidation method or another, and the consolidation method is not adding any new information, the most optimized consolidation method ($h_{kc}^{opt}$) reflects the capacity brought by it:
    \begin{equation}
        \delta_{ds} = P(S{=}1 \mid M, \{h_{id}, h_{ds}, h_{kc}^{opt}\}) - P(S{=}1 \mid M, \{h_{id}\})
    \end{equation}

    \item \textbf{Consolidation Optimization Gap ($\delta_{kc}$)}: The performance difference between the default consolidation and an \textit{optimized} template ($h_{kc}^{opt}$):
    \begin{equation}
        \delta_{kc} = P(S{=}1 \mid M, \{h_{id}, h_{ds}, h_{kc}^{opt}\}) - P(S{=}1 \mid M, \{h_{id}, h_{ds}, h_{kc}^{base}\})
    \end{equation}

    \item \textbf{Application Gap ($\delta_{app}$)}: The residual gap under the most optimized discovery-consolidation pipeline, representing the fundamental execution bottleneck. Note that this application capacity---the ability to reason and plan with acquired knowledge---underlies every stage of the loop, from identification to discovery to consolidation. Therefore it can be regarded as a foundational capability:
    \begin{equation}
        \delta_{app} = 1 - P(S=1 \mid M, \{h_{id}, h_{ds}, h_{kc}^{opt}\})
    \end{equation}
\end{enumerate}

Following this sequential attribution, the total capacity space (100\%) is partitioned as:
\begin{equation}
    1 = P(S=1 \mid M, \emptyset) + \delta_{id} + \delta_{ds} + \delta_{kc} + \delta_{app}
\end{equation}

We further report the ratio of these performance gaps to each model's baseline performance to assess their relative significance. For instance, a 4\% absolute performance gap translates to a $0.15\times$ relative impact for Gemini-3-Pro (baseline 26.0\%), whereas it represents a much more substantial $0.38\times$ impact for the smaller Qwen-3-32B (baseline 10.5\%).

The \textit{relative gap ratio} $r$ for any gap $\delta \in \{\delta_{id}, \delta_{ds}, \delta_{kc}, \delta_{app}\}$ is defined as:
\begin{equation}
    r_{\delta} = \frac{\delta}{P(S=1 \mid M, \emptyset)}
\end{equation}

%% file: appendices/app_limitations.tex
\section{Limitations and Future Work}
\label{app:limitations}

\paragraph{Scaffold specificity.}
All models are evaluated through a single general-purpose code agent scaffold (Claude Code). We chose this design for cross-model comparability and reproducibility, but it means our results reflect the joint performance of model and scaffold rather than an upper bound on what current LLMs can achieve. A purpose-built agent stack with robust memory management, structured experiment tracking, retrieval over prior discoveries, and deliberate tool-use orchestration could plausibly perform better. We encourage the community to use \scicrafter{} as a testbed for evaluating stronger agent architectures; the benchmark's MCP-based interface makes it straightforward to swap in different scaffolds.

\paragraph{Synthetic environment.}
While our Minecraft-based environment isolates core discovery-to-application challenges with clean feedback and deterministic mechanics, it remains synthetic and may not capture all complexities of real-world scientific discovery, such as noisy observations, ambiguous feedback, or open-ended hypothesis spaces. Conclusions drawn from this setting should be viewed as indicative rather than definitive about agent capabilities in broader domains.

\paragraph{Decomposition confounds.}
Our four-way decomposition uses oracle interventions whose effects are not fully orthogonal. For instance, hints may simultaneously identify knowledge gaps and provide metacognitive cues that prompt exploration; the scientist sub-agent's benefit may partly stem from encouraging experimentation at all rather than from its structured template. The measured gaps are therefore best interpreted as the marginal utility of different scaffolding interventions rather than as clean measurements of isolated cognitive capacities.

\paragraph{Future work.}
We plan several extensions in future work. We will incorporate vision input to assess multimodal capabilities. A notable feature of our task design is that it provides paired textual and visual versions of each task, enabling researchers to disentangle multimodal perception from reasoning ability. We also plan to support randomization of the underlying environment dynamics to prevent solutions based on memorization. Moreover, we plan to evaluate additional agent architectures to better separate scaffold effects from model capabilities.

%% file: appendices/app_tasks.tex
\section{Task Specifications}
\label{app:tasks}

This appendix specifies the task contracts and YAML task files used in
\scicrafter. Tasks are organized into five families (A--E), each instantiated
at five difficulty levels (L1--L5). Difficulty does not increase as a purely
quantitative ``more blocks'' scaling: higher levels cross discrete redstone
\emph{mechanism thresholds} (e.g., attenuation and repeater delay semantics),
which forces qualitatively different engineering patterns (e.g., hierarchical
distribution, delay compensation).

\subsection{Common Setup and Contracts}
\label{app:tasks:common}

\paragraph{World.}
All tasks run in a flat, empty creative-mode ``redstone lab'' with a bounded
build region: a radius-$10$ cube centered at an \emph{anchor} position (default
$(0,4,0)$).
The evaluator resets this region before each new attempt.

\paragraph{Component palette.}
To focus evaluation on spatiotemporal reasoning rather than component breadth,
we restrict the palette to ``redstone fundamentals'': \emph{stone button,
redstone wire, redstone repeater, redstone torch, redstone lamp}, plus inert
support blocks (stone, glass) and \texttt{air} for removal.

\paragraph{I/O and tick-level logging.}
Each task provides a single input stimulus (button press) and uses $N$ lamps as
outputs. The evaluator records the on/off state of each lamp at every game tick
(20Hz) and checks a temporal contract $\varphi$ with a $\pm 1$ tick tolerance
(50ms) unless stated otherwise.

\paragraph{Metrics.}
Primary evaluation is functional correctness (Acc: whether $\varphi$ passes).
We additionally track interaction and stability signals, including the number
of environment verification trials (button presses/observations),
Tokens/Attempt, and Engineer rebuild Trials$_{Eng}$.

\paragraph{Difficulty levels.}
Across task families, difficulty levels scale the number of output lamps as
$N \in \{4, 8, 16, 32, 64\}$ for L1--L5 (Table~\ref{tab:taskFamily}), while each
family introduces additional spatial/temporal constraints (e.g., topology
restrictions, delay patterns, distance heterogeneity, pulse duration).

\input{tables/taskFamily}

\subsection{Family A: Simultaneous N-Lamp Control}
\label{app:tasks:familyA}

\paragraph{Goal.}
Build a circuit that activates $N$ lamps (nearly) simultaneously when the
button is pressed.

\paragraph{Why it is challenging.}
For small $N$, a symmetric star/branch pattern can often succeed. As $N$ grows
within a fixed footprint, the agent must (i) distribute power to many
endpoints without unintended dust auto-connections, and (ii) respect signal
attenuation (signal strength decays and vanishes after 15), which eventually
makes repeaters mandatory. Repeaters, however, are not ``free'': they are
directional and introduce quantized delay, so maintaining simultaneity
requires joint reasoning about distance \emph{and} delay.

\paragraph{Level schedule.}
We use the shared $N$ schedule in Table~\ref{tab:taskFamily}.

\paragraph{I/O contract.}
Let $t_{\text{press}}$ be the tick when the button is activated, and $t_i$ be
the first tick when lamp $i$ turns on. The simultaneity contract requires:
\begin{equation}
\forall i, j \in [1, N]: |t_i - t_j| \leq 1 \text{ tick.}
\end{equation}

\paragraph{Example task file (L1).} \mbox{}\\
\begin{lstlisting}[basicstyle=\ttfamily\scriptsize]
task_id: A_simultaneous_lights_L1
family: A
level: L1
task_name: Simultaneous Lights (N=4)
task_description: |
  Build a redstone circuit that turns on 4 lamps with (near) zero skew after a single button press.
  This level is meant to validate basic wiring semantics: strong powering via a support block, dust connectivity, and simple fan-out.
difficulty: beginner

world:
  anchor: [0, 4, 0]
  radius: 10

allowed_blocks:
  - minecraft:stone_button
  - minecraft:redstone_lamp
  - minecraft:redstone_wire
  - minecraft:redstone_torch
  - minecraft:redstone_repeater
  - minecraft:stone
  - minecraft:glass
  - minecraft:air

inputs:
  button:
    type: minecraft:stone_button
    quantity: 1

outputs:
  lamps:
    type: minecraft:redstone_lamp
    quantity: 4

contract:
  type: simultaneity
  tolerance_ticks: 1
  num_outputs: 4

requirements:
  - No block may be placed directly on top of a redstone_lamp (the space above each lamp must remain air).
  - All 4 lamps must turn on within +-1 tick of each other after the button press
  - Construction must fit within the radius-10 build region
  - Use only the allowed block palette

evaluation_metrics:
  functional_correctness:
    description: Verify all lamps activate
    weight: 1.0
    check_method: verify_all_lamps_activated
  simultaneity:
    description: Max skew across lamps <= 1 tick
    weight: 1.0
    check_method: verify_max_skew_within
    params:
      tolerance_ticks: 1

test_cases:
  - name: single_press
    sequence:
      - action: press_button
      - action: check_lamps_on

metadata:
  learning_objectives:
    - strong_power_support_block
    - dust_connectivity
    - simple_fanout
  estimated_difficulty_score: 1.0

\end{lstlisting}

\input{tables/tab_knowledge_gaps}

\subsection{Family B: Branch Reach (T-Junction Expansion)}
\label{app:tasks:familyB}

\paragraph{Goal.}
Reach $N$ off-axis lamps from a central trunk using T-junction branches, while
maintaining valid powering (attenuation-aware) and near-simultaneous
activation.

\paragraph{Why it is challenging.}
Unlike Family~A, which primarily rewards symmetry, Family~B constrains the
\emph{wiring topology}: solutions must implement a trunk-and-branch routing
pattern. This exposes different failure modes: long trunk lines need repeater
placement that respects attenuation, while crowded branches are prone to dust
auto-connections that create unintended shortcuts.

\paragraph{Level schedule.}
We use the shared $N$ schedule in Table~\ref{tab:taskFamily}. The maximum
required reach (measured in wire path length) increases from 8 to 20 blocks
across levels.

\paragraph{I/O contract.}
Same simultaneity requirement as Family~A ($\pm$1 tick), with an additional
topology constraint that the solution must contain at least one explicit
T-junction branch (fan-out node) in the redstone dust graph.

\paragraph{Example task file (L1).} \mbox{}\\
\begin{lstlisting}[basicstyle=\ttfamily\scriptsize]
task_id: B_branch_reach_L1
family: B
level: L1
task_name: Branch Reach (T-junction) (N=4)
task_description: |
  Build a trunk-and-branch wiring layout that reaches 4 lamps that are not collinear with the button.
  The circuit must use at least one explicit T-junction and still satisfy the same +/-1 tick simultaneity tolerance.
difficulty: beginner

world:
  anchor: [0, 4, 0]
  radius: 10

allowed_blocks:
  - minecraft:stone_button
  - minecraft:redstone_lamp
  - minecraft:redstone_wire
  - minecraft:redstone_torch
  - minecraft:redstone_repeater
  - minecraft:stone
  - minecraft:glass
  - minecraft:air

inputs:
  button:
    type: minecraft:stone_button
    quantity: 1

outputs:
  lamps:
    type: minecraft:redstone_lamp
    quantity: 4

contract:
  type: branch_reach
  tolerance_ticks: 1
  num_outputs: 4
  max_reach_blocks: 8
  require_t_junction: true

requirements:
  - No block may be placed directly on top of a redstone_lamp (the space above each lamp must remain air).
  - All 4 lamps must turn on within +/-1 tick of each other after the button press
  - At least one T-junction (fan-out) must be present in the wiring topology
  - The solution must respect redstone attenuation (no dead outputs)
  - Construction must fit within the radius-10 build region

evaluation_metrics:
  functional_correctness:
    description: Verify all lamps activate
    weight: 1.0
    check_method: verify_all_lamps_activated
  simultaneity:
    description: Max skew across lamps <= 1 tick
    weight: 1.0
    check_method: verify_max_skew_within
    params:
      tolerance_ticks: 1

test_cases:
  - name: single_press
    sequence:
      - action: press_button
      - action: check_lamps_on

metadata:
  learning_objectives:
    - t_junction_branching
    - off_axis_routing
    - attenuation_awareness
  estimated_difficulty_score: 1.0
\end{lstlisting}

\subsection{Family C: Sequential Activation (Delay Pattern)}
\label{app:tasks:familyC}

\paragraph{Goal.}
Activate lamps in a specified temporal order with target inter-stage delays.

\paragraph{Why it is challenging.}
This family tests \emph{temporal synthesis}: the agent must realize a target
delay sequence using repeaters with discrete delay settings (1--4 ticks), while
preventing accidental bypasses (e.g., dust auto-connections) that would trigger
later stages early.

\paragraph{Level schedule.}
We use the shared $N$ schedule in Table~\ref{tab:taskFamily}. Each YAML file
specifies an inter-stage delay vector $\boldsymbol\delta$ of length $N-1$
(e.g., an alternating pattern), which the agent must realize with quantized
repeater delays.

\paragraph{I/O contract.}
Let $\delta_i$ be the specified delay between stage $i$ and stage $i+1$.
The contract requires:
\begin{equation}
\forall i \in [1, N-1]: |t_{i+1} - t_i - \delta_i| \leq 1 \text{ tick.}
\end{equation}

\paragraph{Example task file (L1).}\mbox{}\\

\begin{lstlisting}[basicstyle=\ttfamily\scriptsize]
task_id: C_sequential_activation_L1
family: C
level: L1
task_name: Sequential Activation (N=4, delays=[1,2,1])
task_description: |
  After a button press, activate 4 lamps in order with inter-stage delays [1, 2, 1] ticks (+/-1 tick tolerance).
  This level checks that the agent can realize a basic multi-stage delay line using repeaters.
difficulty: beginner

world:
  anchor: [0, 4, 0]
  radius: 10

allowed_blocks:
  - minecraft:stone_button
  - minecraft:redstone_lamp
  - minecraft:redstone_wire
  - minecraft:redstone_torch
  - minecraft:redstone_repeater
  - minecraft:stone
  - minecraft:glass
  - minecraft:air

inputs:
  button:
    type: minecraft:stone_button
    quantity: 1

outputs:
  lamps:
    type: minecraft:redstone_lamp
    quantity: 4

contract:
  type: sequential
  tolerance_ticks: 1
  delays_ticks: [1, 2, 1]

requirements:
  - No block may be placed directly on top of a redstone_lamp (the space above each lamp must remain air).
  - Lamps must activate in order with the specified inter-stage delays

evaluation_metrics:
  functional_correctness:
    description: Verify order and timing per delays
    weight: 1.0
    check_method: verify_sequence_with_delays
    params:
      delays_ticks: [1, 2, 1]
      tolerance_ticks: 1

test_cases:
  - name: single_press
    sequence:
      - action: press_button
      - action: check_sequence
        params:
          delays_ticks: [1, 2, 1]
          tolerance_ticks: 1

metadata:
  learning_objectives:
    - repeater_delay_basic
    - ordered_activation
  estimated_difficulty_score: 1.0
\end{lstlisting}

\subsection{Family D: Equal-Delay Distribution}
\label{app:tasks:familyD}

\paragraph{Goal.}
Deliver a signal to lamps placed at heterogeneous distances, while ensuring all
lamps still activate (nearly) simultaneously by compensating shorter paths with
additional delay.

\paragraph{Why it is challenging.}
Family~D stresses \emph{delay alignment} under heterogeneous geometry.
Some outputs are inherently ``slow'' because they require longer routes and/or
mandatory repeaters for signal regeneration. To synchronize all outputs, the
agent must intentionally \emph{slow down the fast paths} by inserting
compensation repeaters, turning repeaters into timing buffers rather than just
range extenders.

\paragraph{Level schedule.}
We use the shared $N$ schedule in Table~\ref{tab:taskFamily}. Lamps are
partitioned into distance buckets to induce heterogeneous inherent delays; the
agent must compensate shorter paths so that all outputs satisfy the same
simultaneity tolerance.

\paragraph{I/O contract.}
Despite different path lengths, all lamps must activate simultaneously within
tolerance:
\begin{equation}
\forall i, j \in [1, N]: |t_i - t_j| \leq 1 \text{ tick.}
\end{equation}

\paragraph{Example task file (L1).}\mbox{}\\
\begin{lstlisting}[basicstyle=\ttfamily\scriptsize]
task_id: D_equal_delay_distribution_L1
family: D
level: L1
task_name: Equal-Delay Distribution (N=4, distances={4,8,12,16})
task_description: |
  Place and wire a circuit so that 4 lamps at different distances still turn on within +/-1 tick of each other after a button press.
  The intended solution uses repeaters as compensatory delay elements on the faster (shorter) paths.
difficulty: beginner

world:
  anchor: [0, 4, 0]
  radius: 10

allowed_blocks:
  - minecraft:stone_button
  - minecraft:redstone_lamp
  - minecraft:redstone_wire
  - minecraft:redstone_torch
  - minecraft:redstone_repeater
  - minecraft:stone
  - minecraft:glass
  - minecraft:air

inputs:
  button:
    type: minecraft:stone_button
    quantity: 1

outputs:
  lamps:
    type: minecraft:redstone_lamp
    quantity: 4

contract:
  type: equal_delay
  tolerance_ticks: 1
  target_distances: [4, 8, 12, 16]

requirements:
  - No block may be placed directly on top of a redstone_lamp (the space above each lamp must remain air).
  - All 4 lamps must turn on within +/-1 tick (skew <= 1)
  - The solution must compensate different distances using repeater delays

evaluation_metrics:
  functional_correctness:
    description: Verify all lamps activate
    weight: 1.0
    check_method: verify_all_lamps_activated
  skew_tolerance:
    description: Max skew across lamps <= 1 tick
    weight: 1.0
    check_method: verify_max_skew_within
    params:
      tolerance_ticks: 1

test_cases:
  - name: single_press
    sequence:
      - action: press_button
      - action: check_lamps_on

metadata:
  learning_objectives:
    - delay_compensation_basic
    - repeater_as_buffer
  estimated_difficulty_score: 1.0
\end{lstlisting}

\subsection{Family E: Pulse Shaping (Fixed On-Duration)}
\label{app:tasks:familyE}

\paragraph{Goal.}
On each button press, keep all $N$ lamps lit for a target duration $\tau$
ticks, then return to off.

\paragraph{Why it is challenging.}
The input button produces a pulse width determined by the game mechanics.
Family~E requires shaping this pulse into a target on-duration $\tau$, then
distributing the resulting waveform to $N$ outputs without introducing large
skew. This typically requires a pulse shaper (e.g., monostable / edge-triggered
variants realizable with torches and repeaters), not just copying the input
wire.

\paragraph{Level schedule.}
We use the shared $N$ schedule in Table~\ref{tab:taskFamily}. The target
duration increases from $\tau=4$ to $\tau=12$ ticks across levels.

\paragraph{I/O contract.}
Let $t_{\text{press}}$ be the button press time. For each lamp $i$, define
$t_i^{\text{on}}$ as the first on tick and $t_i^{\text{off}}$ as the first off
tick after activation. The contract requires:
\begin{align}
t_i^{\text{on}} &\in [t_{\text{press}},\, t_{\text{press}} + 1], \\
t_i^{\text{off}} &\in [t_{\text{press}} + \tau - 1,\, t_{\text{press}} + \tau + 1].
\end{align}

\paragraph{YAML note.}
For compactness, this appendix includes YAML examples for Families A--D.
Family~E follows the same schema and is released with the benchmark harness.

\subsection{Task File Format (YAML)}
\label{app:tasks:format}

Each task is specified in YAML and stored under \texttt{appendices/task\_specs/}.
The released files follow the same high-level structure:

\begin{lstlisting}
# Identity
task_id: <string>
family: <A|B|C|D|E>
level: <L1|L2|L3|L4|L5>
task_name: <string>
task_description: <string>

# Build region and palette
world:
  anchor: [x, y, z]
  radius: 10
allowed_blocks:
  - minecraft:stone_button
  - minecraft:redstone_wire
  - minecraft:redstone_repeater
  - minecraft:redstone_torch
  - minecraft:redstone_lamp
  - minecraft:stone
  - minecraft:glass
  - minecraft:air

# I/O specification
inputs:  { ... }      # one stone button
outputs: { ... }      # N lamps
contract: { ... }     # family-specific parameters (e.g., tolerance, delays, tau)

# Test harness
test_cases:
  - name: <string>
    sequence:
      - {action: press_button}
      - {action: check_<...>, params: {...}}

# Optional metadata (not used by the checker)
metadata:
  learning_objectives: [ ... ]
\end{lstlisting}

\input{tables/tab_metrics}

%% file: tables/taskFamily.tex

\newcolumntype{L}{>{\raggedright\arraybackslash}X}

\begin{table*}[t!]
\footnotesize
\begin{tabularx}{\textwidth}{@{} l L L L L L @{}} 
\toprule
\textbf{Level} & \textbf{Family A (Simultaneous)} & \textbf{Family B (Branch Reach)} & \textbf{Family C (Sequential)} & \textbf{Family D (Equal Delay)} & \textbf{Family E (Pulse)} \\
\midrule
\textbf{L1} & $N$=4, skew $\le$1 tick, radius=10 & $N$=4, max reach=8, skew $\le$1 tick, T-junction & $N$=4, delays=[1,2,1], tol=$\pm$1 tick & $N$=4, distance buckets=\{4,8,12,16\}, skew $\le$1 tick & $N$=4, $\tau$=4 ticks \\
\textbf{L2} & $N$=8, skew $\le$1 tick, radius=10 & $N$=8, max reach=12, skew $\le$1 tick, T-junction & $N$=8, delays=[1,2]$\times$3+[1], tol=$\pm$1 tick & $N$=8, distance buckets=\{4,8,12,16\}, skew $\le$1 tick & $N$=8, $\tau$=6 ticks \\
\textbf{L3} & $N$=16, skew $\le$1 tick, radius=10 & $N$=16, max reach=15, skew $\le$1 tick, T-junction & $N$=16, delays=[1,2]$\times$7+[1], tol=$\pm$1 tick & $N$=16, distance buckets=\{4,8,12,16\}, skew $\le$1 tick & $N$=16, $\tau$=8 ticks \\
\textbf{L4} & $N$=32, skew $\le$1 tick, radius=10 & $N$=32, max reach=18, skew $\le$1 tick, \textit{repeaters required} & $N$=32, delays=[1,2]$\times$15+[1], tol=$\pm$1 tick & $N$=32, distance buckets=\{4,8,12,16\}, skew $\le$1 tick & $N$=32, $\tau$=10 ticks \\
\textbf{L5} & $N$=64, skew $\le$1 tick, radius=10 & $N$=64, max reach=20, skew $\le$1 tick, \textit{repeaters required} & $N$=64, delays=[1,2]$\times$31+[1], tol=$\pm$1 tick & $N$=64, distance buckets=\{4,8,12,16\}, skew $\le$1 tick & $N$=64, $\tau$=12 ticks \\
\bottomrule
\end{tabularx}
\caption{Task families and level parameters. All families scale the number of output lamps as $N \in \{4,8,16,32,64\}$ for L1--L5, while introducing family-specific spatial/temporal constraints.}
\label{tab:taskFamily}
\end{table*}

%% file: tables/tab_knowledge_gaps.tex
\begin{table*}[t]
  \centering
  \footnotesize
  \label{tab:tasks}
  \resizebox{\textwidth}{!}{%
  \begin{tabular}{c c p{0.68\textwidth} p{0.18\textwidth}}
  \toprule
  \textbf{Level} & \textbf{N} & \textbf{Knowledge Gap} & \textbf{Hint} \\
  \midrule
  L1 & 4 & Button powering primitives: a button must strongly power a supporting block; dust only conducts via cardinal (N/S/E/W) adjacency; lamps require adjacent powered dust. & Strong-Power-Support-Block \\
  L2 & 8 & Nested fanout topology (hub $\rightarrow$ branch hubs) to reach off-axis/diagonal lamps without diagonal dust connectivity; preserve symmetry to avoid accidental skew. & Nested-Hub-Fanout \\
  L3 & 16 & Signal strength attenuates with wire distance; multi-radius branching (e.g., two rings) must be planned so all lamps receive nonzero power within the dust budget. & Signal-Strength-Decay \\
  L4 & 32 & High-fanout distribution under radius and attenuation constraints: shared backbones/rails and dense symmetric branching; avoid unintended dust auto-connections/shorts in crowded layouts. & Attenuation-Aware-Fanout \\
  L5 & 64 & Repeater semantics for scaling: where to insert repeaters to regenerate signal past the attenuation limit; repeaters are directional and add delay, so boosters must be placed symmetrically to maintain $\pm$1 tick simultaneity. & Repeater-Signal-Regeneration \\
  \bottomrule
  \end{tabular}}

  \label{tab:knowledge_gap_breakdown}
  \caption{\textbf{Breakdown of knowledge gaps per level for Task A}. This task variant requires activating $N$ lamps simultaneously. \textbf{Knowledge Gap} lists the environment knowledge agents must discover to design a working solution. \textbf{Hint} is a concise directional keyword indicating the missing area without revealing the specific mechanism.}
\end{table*}

%% file: tables/tab_metrics.tex
\begin{table}[b]
\centering
\footnotesize
\begin{tabularx}{\linewidth}{@{} l X @{}}
\toprule
\textbf{Metric} & \textbf{Definition} \\
\midrule
Acc &
Fraction of tasks whose constructed device satisfies the spatiotemporal
contract $\varphi$ over all evaluator test cases. \\
Tokens/Attempt &
Total LLM tokens consumed per task attempt (including planning and tool calls),
averaged over attempts. \\
Trials$_{Eng}$ &
Number of build--revise cycles executed by the Engineer within a task before
final submission (lower is better). \\
Verification Trials &
Number of environment interaction trials used for exploration and validation
(e.g., button press plus subsequent observations), subject to a fixed budget. \\
\bottomrule
\end{tabularx}
\caption{Evaluation metrics tracked for \scicrafter.}
\label{tab:metrics}
\end{table}

%% file: appendices/app_exp_details.tex
\section{Experiment Details}
\label{app:exp_details}

\paragraph{Code Agent Framework}
We evaluate all models using a state-of-the-art code agent framework~\citep{dong2025survey,google2025geminicodeassist,anthropic2025claudecode}. Our tasks require sustained interaction and complex reasoning, making the code agent paradigm well-suited as an evaluation vehicle: it is designed to interface with complex environments and construct functional artifacts through executable code~\citep{soni2025coding, wang2024executable}, providing a faithful representation of current frontier AI capabilities. Specifically, we use Claude Code~\citep{anthropic2025claudecode}, though our method generalizes to other code agents.

\paragraph{Model Selection}
Our selection spans closed- and open-source models, frontier reasoning systems, and various parameter scales within model families: GPT-5.2, Claude-Opus-4.5, Gemini-3-Pro, Grok-4, GLM-4.7, Qwen3-235B-MoE, Qwen2.5-72B-Instruct, and Qwen3-32B. Models are evaluated by swapping the underlying API while keeping all other components fixed.

\paragraph{Verification Trials}
We enforce a budget of 50 verification trials per task as a normalized compute metric that sidesteps token-count variability across model architectures. A verification trial consists of the agent triggering a mechanism (e.g., pressing a button) and observing subsequent block-state transitions. Importantly, the environment returns only raw state changes, not success signals from the evaluation script; agents must autonomously interpret these responses to assess functionality. All trials conducted by the scientist sub-agent count toward this global budget, forcing agents to trade off between exploration and final validation.

\paragraph{Evaluation Protocol}
Models are evaluated in a curriculum setting: agents progress from L1 to L5, carrying forward accumulated experience. The primary metric is success rate (percentage of tasks completed), averaged over eight independent runs. Results for independent (non-curriculum) evaluation are reported in Section~\ref{app:more_exp_results}.

%% file: appendices/app_more_exp_results.tex
\clearpage
\section{More Experiment Results}
\label{app:more_exp_results}

We report more experiment results here. 

In Table~\ref{tab:main_results_indep_wsci}, we report results under the Independent setting, where tasks are solved independently rather than in curriculum order (L1 to L5 for each task). While the main paper centers on the Curriculum setting—which better reflects models' upper-bound capacity—we include Independent results to examine how models perform without progressive scaffolding. As the results show, the performance under the independent setting is generally lower than under the curriculum setting, indicating that progressive task exposure helps build knowledge that transfers to harder challenges.

We also reports results with the scientist sub-agent alone (using the optimized knowledge consolidation structure) without hints. The main paper reports the performance difference between models with hints and models with hints plus the scientist sub-agent, ensuring all gaps sum to 100\%. Here, we isolate the contribution of the scientist sub-agent alone. The results show that it alone attains a comparable improvement to combining it with hints. For example Gemini-3-Pro improved 9.0\% from baseline to w/scientist sub-agent, which is 2.5\% smaller than the 11.5\% improvement from w/hint to w/hint+scientist. This indicates that oracle hints can amplify the effectiveness of the scientist sub-agent.

Tables~\ref{tab:breakdown_baseline} through~\ref{tab:breakdown_full} provide a detailed breakdown of Gemini-3-Pro's performance by task type and difficulty level. We report results across four conditions—baseline, with hints, with scientist alone, and with hints plus scientist—under both Independent and Curriculum settings. The baseline model fails completely at L4, but with hints or the scientist sub-agent, models begin to tackle L4 challenges. However, none succeed at L5, leaving this as an open challenge for future models. We also observe that Task C is consistently the most difficult, achieving the lowest scores across all conditions. This may be because its sequential requirements pose greater demands on precise spatial-temporal reasoning—a notable weakness of current language models.

\input{tables/tab_main_results_complete}
\input{tables/tab_per_level_breakdown}

\clearpage

%% file: tables/tab_main_results_complete.tex
\begin{table*}[t]
    \centering
    \footnotesize
    \setlength{\tabcolsep}{2.5pt}
    \renewcommand{\arraystretch}{1.05}

    \begin{tabular}{l c c c c c c c c}
        \toprule
        & \multicolumn{4}{c}{\textbf{Independent.}}
        & \multicolumn{4}{c}{\textbf{Curriculum.}} \\
        \cmidrule(lr){2-5} \cmidrule(lr){6-9}
        \textbf{Model}
        & \makecell[c]{baseline} & \makecell[c]{w/ hint} & \makecell[c]{w/ sci.\\sub-agent} & \makecell[c]{w/ hint\\+ scientist}
        & \makecell[c]{baseline} & \makecell[c]{w/ hint} & \makecell[c]{w/ sci.\\sub-agent} & \makecell[c]{w/ hint\\+ scientist} \\
        \midrule
        \texttt{gemini-3-pro}       & $15.5$ & $43.0$ & $34.5$ & $56.5$ & $26.0$ & $52.5$ & $35.0$ & $64.0$  \\
        \texttt{gpt-5.2}            & $22.0$ & $40.5$ & $33.5$ & $63.0$ & $25.5$ & $51.0$ & $32.5$ & $60.0$  \\
        \texttt{claude-opus-4.5}    & $16.5$ & $44.0$ & $32.0$ & $56.5$ & $21.0$ & $46.0$ & $36.0$ & $59.0$  \\
        \texttt{glm-4.7}            & $16.0$ & $40.5$ & $23.5$ & $50.5$ & $23.0$ & $45.5$ & $33.5$ & $53.0$  \\
        \texttt{grok-4}             & $13.5$ & $34.5$ & $25.0$ & $50.5$ & $22.5$ & $42.5$ & $33.5$ & $56.5$  \\
        \texttt{qwen3-235b}         & $10.0$ & $37.0$ & $25.5$ & $52.0$ & $18.5$ & $42.5$ & $25.5$ & $55.5$  \\
        \texttt{qwen2.5-72b}        & $11.5$ & $37.0$ & $25.0$ & $39.5$ & $14.0$ & $29.0$ & $30.0$ & $43.0$  \\
        \texttt{qwen3-32b}          & $11.0$ & $30.5$ & $18.5$ & $41.5$ & $10.5$ & $37.5$ & $20.0$ & $46.5$  \\
        \bottomrule
    \end{tabular}

    \caption{\textbf{Main Performance + Ablations.}
    Success rates (\%) aggregated across all 25 tasks over 8 runs ($k/200 \times 100$), for \textbf{Independent and Curriculum settings}. \textbf{Hints} guide agents toward the correct inquiry, revealing the performance gap in ``asking the right questions''. The \textbf{scientist sub-agent} enhances the ability to do systematic control experiments, highlighting gaps in robust knowledge discovery. All experiments use a 50-trial budget.}
    \label{tab:main_results_indep_wsci}
\end{table*}

%% file: tables/tab_per_level_breakdown.tex
\begin{table*}[t]
\centering
\scriptsize
\setlength{\tabcolsep}{1.5pt}
\begin{tabular*}{\textwidth}{@{\extracolsep{\fill}} l | ccccc c | ccccc c @{}}
\toprule
\textbf{Level} & \multicolumn{6}{c|}{\textbf{Independent Setting (Baseline)}} & \multicolumn{6}{c}{\textbf{Curriculum Setting (Baseline)}} \\
\cmidrule(lr){2-7} \cmidrule(lr){8-13}
\textbf{Description} & \textbf{Task A} & \textbf{Task B} & \textbf{Task C} & \textbf{Task D} & \textbf{Task E} & \textbf{Avg} & \textbf{Task A} & \textbf{Task B} & \textbf{Task C} & \textbf{Task D} & \textbf{Task E} & \textbf{Avg} \\
\midrule
L1 (Primitive)    & $25.0$ & $50.0$ & $37.5$ & $37.5$ & $37.5$ & \textbf{37.5} & $75.0$ & $75.0$ & $50.0$ & $50.0$ & $37.5$ & \textbf{57.5} \\
L2 (Basic)        & $25.0$ & $25.0$ & $25.0$ & $25.0$ & $50.0$ & \textbf{30.0} & $50.0$ & $37.5$ & $0.0$  & $37.5$ & $37.5$ & \textbf{32.5} \\
L3 (Intermediate) & $12.5$ & $0.0$  & $12.5$ & $0.0$  & $12.5$ & \textbf{7.5}  & $37.5$ & $37.5$ & $12.5$ & $12.5$ & $37.5$ & \textbf{27.5} \\
L4 (Advanced)     & $0.0$  & $0.0$  & $0.0$  & $0.0$  & $12.5$ & \textbf{2.5}  & $25.0$ & $12.5$ & $0.0$  & $0.0$  & $25.0$ & \textbf{12.5} \\
L5 (Complex)      & $0.0$  & $0.0$  & $0.0$  & $0.0$  & $0.0$  & \textbf{0.0}  & $0.0$  & $0.0$  & $0.0$  & $0.0$  & $0.0$  & \textbf{0.0} \\
\bottomrule
\end{tabular*}
\caption{\textbf{Baseline Performance Breakdown.} Per-task success rates (\%, $k/8 \times 100$) for Gemini-3-Pro without any assistance.}
\label{tab:breakdown_baseline}
\end{table*}

\begin{table*}[t]
\centering
\scriptsize
\setlength{\tabcolsep}{1.5pt}
\begin{tabular*}{\textwidth}{@{\extracolsep{\fill}} l | ccccc c | ccccc c @{}}
\toprule
\textbf{Level} & \multicolumn{6}{c|}{\textbf{Independent Setting (w/ Hint)}} & \multicolumn{6}{c}{\textbf{Curriculum Setting (w/ Hint)}} \\
\cmidrule(lr){2-7} \cmidrule(lr){8-13}
\textbf{Description} & \textbf{Task A} & \textbf{Task B} & \textbf{Task C} & \textbf{Task D} & \textbf{Task E} & \textbf{Avg} & \textbf{Task A} & \textbf{Task B} & \textbf{Task C} & \textbf{Task D} & \textbf{Task E} & \textbf{Avg} \\
\midrule
L1 (Primitive)    & $100.0$ & $62.5$ & $50.0$ & $75.0$ & $87.5$ & \textbf{75.0} & $75.0$ & $87.5$ & $62.5$ & $87.5$ & $100.0$ & \textbf{82.5} \\
L2 (Basic)        & $87.5$ & $62.5$ & $50.0$ & $62.5$ & $50.0$ & \textbf{62.5} & $87.5$ & $37.5$ & $87.5$ & $75.0$ & $75.0$ & \textbf{72.5} \\
L3 (Intermediate) & $75.0$ & $37.5$ & $37.5$ & $87.5$ & $25.0$ & \textbf{52.5} & $75.0$ & $62.5$ & $50.0$ & $87.5$ & $62.5$ & \textbf{67.5} \\
L4 (Advanced)     & $50.0$ & $37.5$ & $0.0$  & $37.5$ & $0.0$  & \textbf{25.0} & $37.5$ & $25.0$ & $50.0$ & $62.5$ & $25.0$ & \textbf{40.0} \\
L5 (Complex)      & $0.0$  & $0.0$  & $0.0$  & $0.0$  & $0.0$  & \textbf{0.0}  & $0.0$  & $0.0$  & $0.0$  & $0.0$  & $0.0$  & \textbf{0.0} \\
\bottomrule
\end{tabular*}
\caption{\textbf{Augmented with Hint.} Breakdown of performance when the agent is provided with discovery target hints.}
\label{tab:breakdown_hint}
\end{table*}

\begin{table*}[t]
\centering
\scriptsize
\setlength{\tabcolsep}{1.5pt}
\begin{tabular*}{\textwidth}{@{\extracolsep{\fill}} l | ccccc c | ccccc c @{}}
\toprule
\textbf{Level} & \multicolumn{6}{c|}{\textbf{Independent Setting (w/ Scientist)}} & \multicolumn{6}{c}{\textbf{Curriculum Setting (w/ Scientist)}} \\
\cmidrule(lr){2-7} \cmidrule(lr){8-13}
\textbf{Description} & \textbf{Task A} & \textbf{Task B} & \textbf{Task C} & \textbf{Task D} & \textbf{Task E} & \textbf{Avg} & \textbf{Task A} & \textbf{Task B} & \textbf{Task C} & \textbf{Task D} & \textbf{Task E} & \textbf{Avg} \\
\midrule
L1 (Primitive)    & $87.5$ & $62.5$ & $62.5$ & $75.0$ & $62.5$ & \textbf{70.0} & $62.5$ & $62.5$ & $62.5$ & $50.0$ & $50.0$ & \textbf{57.5} \\
L2 (Basic)        & $37.5$ & $50.0$ & $62.5$ & $62.5$ & $50.0$ & \textbf{52.5} & $62.5$ & $37.5$ & $37.5$ & $37.5$ & $75.0$ & \textbf{50.0} \\
L3 (Intermediate) & $50.0$ & $25.0$ & $37.5$ & $25.0$ & $37.5$ & \textbf{35.0} & $50.0$ & $50.0$ & $50.0$ & $37.5$ & $50.0$ & \textbf{47.5} \\
L4 (Advanced)     & $12.5$ & $12.5$ & $12.5$ & $0.0$  & $37.5$ & \textbf{15.0} & $25.0$ & $25.0$ & $12.5$ & $12.5$ & $25.0$ & \textbf{20.0} \\
L5 (Complex)      & $0.0$  & $0.0$  & $0.0$  & $0.0$  & $0.0$  & \textbf{0.0}  & $0.0$  & $0.0$  & $0.0$  & $0.0$  & $0.0$  & \textbf{0.0} \\
\bottomrule
\end{tabular*}
\caption{\textbf{Augmented with Scientist Sub-agent.} Breakdown of performance when the agent is augmented with the Scientist sub-agent.}
\label{tab:breakdown_scientist}
\end{table*}

\begin{table*}[t]
\centering
\scriptsize
\setlength{\tabcolsep}{1.5pt}
\begin{tabular*}{\textwidth}{@{\extracolsep{\fill}} l | ccccc c | ccccc c @{}}
\toprule
\textbf{Level} & \multicolumn{6}{c|}{\textbf{Independent Setting (w/ Hint + Scientist)}} & \multicolumn{6}{c}{\textbf{Curriculum Setting (w/ Hint + Scientist)}} \\
\cmidrule(lr){2-7} \cmidrule(lr){8-13}
\textbf{Description} & \textbf{Task A} & \textbf{Task B} & \textbf{Task C} & \textbf{Task D} & \textbf{Task E} & \textbf{Avg} & \textbf{Task A} & \textbf{Task B} & \textbf{Task C} & \textbf{Task D} & \textbf{Task E} & \textbf{Avg} \\
\midrule
L1 (Primitive)    & $75.0$ & $50.0$ & $50.0$ & $100.0$ & $87.5$ & \textbf{72.5} & $87.5$ & $100.0$ & $75.0$ & $100.0$ & $100.0$ & \textbf{92.5} \\
L2 (Basic)        & $100.0$ & $100.0$ & $87.5$ & $87.5$ & $87.5$ & \textbf{92.5} & $75.0$ & $87.5$ & $87.5$ & $100.0$ & $100.0$ & \textbf{90.0} \\
L3 (Intermediate) & $62.5$ & $62.5$ & $50.0$ & $87.5$ & $87.5$ & \textbf{70.0} & $87.5$ & $87.5$ & $62.5$ & $75.0$ & $100.0$ & \textbf{82.5} \\
L4 (Advanced)     & $37.5$ & $62.5$ & $62.5$ & $12.5$ & $62.5$ & \textbf{47.5} & $50.0$ & $37.5$ & $75.0$ & $50.0$ & $62.5$ & \textbf{55.0} \\
L5 (Complex)      & $0.0$  & $0.0$  & $0.0$  & $0.0$  & $0.0$  & \textbf{0.0}  & $0.0$  & $0.0$  & $0.0$  & $0.0$  & $0.0$  & \textbf{0.0} \\
\bottomrule
\end{tabular*}
\caption{\textbf{Full Method (Hint + Scientist).} Breakdown of performance with both augmentations.}
\label{tab:breakdown_full}
\end{table*}

%% file: appendices/app_env.tex
\section{Env Specifications}
\label{app:env}

\paragraph{Environment Setup}
In our benchmark, the Minecraft environment is set into \textit{creative} mode where block items are provided to agents so that we can focus on building instead of collecting resources.  

To focus evaluation on spatiotemporal-causal reasoning rather than
component breadth, this version restricts to common redstone components:
\emph{stone button, redstone wire, redstone repeater, redstone torch,
redstone lamp} (with inert supports such as stone and glass).
More advanced components (e.g., comparators, observers) are excluded and
reserved for future extensions.

\paragraph{MCP Calls}
To facilitate the community to test our tasks, we wrap the interaction interface between the environment and the agent using  \textbf{Model Context Protocol (MCP)}\footnote{\url{https://modelcontextprotocol.io}}. MCP has become the standard function call protocol for LLMs so the capability of LLMs would not be affected by the function call format. 

Specifically, our observation MCP calls include \texttt{get-block-state} to capture desired block state,  
\texttt{get-event-stream} to receive recent event sequence, and \texttt{scan-redstone-area} 
to return nearby redstone components. To configure blocks, the calls include  \texttt{set-block} to any type, position, or state, thereby
capturing all building operations.  
The \texttt{activate-button} action
automatically locates and toggles the device's button.

%% file: appendices/app_prompt_templates.tex
\section{Prompt and Experiment Templates}
\label{app:prompt_templates}

This appendix provides full prompt templates used by the Scientist and Engineer
agents, along with the experiment write-up template used for controlled
game-mechanics exploration.

\subsection{Scientist Agent Prompt Template (YAML)}
\begin{lstlisting}
# Scientist Agent Prompt Template
# Purpose: Discover environment laws through controlled experimentation
# Based on: SciCraft paper Section 4.1 (Scientist Agent)

agent_name: Scientist Agent
agent_role: Environment Law Discovery and Documentation
version: 1.0

# System prompt - describes the agent's role and capabilities
system_prompt: |
  You are the Scientist Agent in the Minecraft framework. Your role is to discover
  fundamental environment laws through controlled experimentation in Minecraft.

  Core Principle: **NEVER GUESS**. All laws must be validated through reproducible experiments
  with minimum support count >= 3.

question: |
  {question_from_engineer}

# Main task prompt with placeholders
experiment_template: |
  {experiment_template}

knowledge_book: |
  {knowledge_book}
\end{lstlisting}

\subsection{Engineer Agent Prompt Template (YAML)}
\begin{lstlisting}
# Engineer Agent Prompt Template
# Purpose: Generate construction plans for functional devices in Minecraft
# Based on: SciCraft paper Section 4.2 (Engineer Agent)

agent_name: Engineer Agent
agent_role: Device Construction and Task Execution
version: 1.0

# System prompt - describes the agent's role and capabilities
system_prompt: |
  You are the Engineer Agent in the  Minecraft framework. Your role is to construct
  functional redstone devices in Minecraft that satisfy precise I/O specifications.

  You must NEVER guess about environment laws. When uncertain, delegate to the Scientist Agent.

# Main task prompt with placeholders
task_prompt_template: |
  # TASK SPECIFICATION

  {task_specification}

  ---

  # AVAILABLE COMPONENTS

  You may use the following Minecraft blocks:
  {available_blocks}

  ---

  # Knowledge Book

  The following laws have been discovered and validated through experiments:

  {knowledge_book}
\end{lstlisting}

\subsection{Game Mechanics Exploration Experiment Template (Markdown)}
\begin{lstlisting}
# Game Mechanics Exploration Experiment Template

## Experiment Info
**Experiment #**: ________  
**Date**: ________  
**Experimenter**: ________

---

## 1. Research Question
What do I want to figure out?

_[e.g., Under what conditions will a device activate? How do two components interact?]_

---

## 2. My Hypothesis
Based on my experience, what do I think will happen?

_[Write down your prediction]_

---

## 3. Experiment Design

### Variables to Test
**What I will change** (Independent Variable): ________

**What I will observe** (Dependent Variable): ________

**What needs to stay constant** (Control Variables):
- ________
- ________
- ________

### Control Setup
- **Control Group**: ________ (baseline test without changes)
- **Experimental Group**: ________ (test with changes)

---

## 4. Experiment Steps

### Preparation
1. ________
2. ________
3. ________

### Testing Process
**Step 1**: ________  
-> Observation: ________

**Step 2**: ________  
-> Observation: ________

**Step 3**: ________  
-> Observation: ________

(Add more steps as needed)

---

## 5. Experiment Record

### Data Recording Table

| Trial # | Changed Condition | Observed Result | Matches Prediction? | Notes |
|---------|------------------|-----------------|-------------------|-------|
| 1       |                  |                 | [ ] Yes [ ] No    |       |
| 2       |                  |                 | [ ] Yes [ ] No    |       |
| 3       |                  |                 | [ ] Yes [ ] No    |       |
| 4       |                  |                 | [ ] Yes [ ] No    |       |
| 5       |                  |                 | [ ] Yes [ ] No    |       |

### Detailed Observations
_[Record any interesting phenomena, unexpected situations, details]_

### Screenshots/Diagrams
[Record diagrams of key configurations or phenomena here]

---

## 6. Experiment Results

### What did I discover?
_[Objectively describe the observed phenomena]_

### Was my hypothesis correct?
[ ] Completely correct  
[ ] Partially correct  
[ ] Completely incorrect

**Explanation**: ________

---

## 7. Analysis & Summary

### Why did this happen?
_[Try to explain the underlying mechanism or pattern]_

### What's the practical application?
_[Real-world use cases]_

### What's still uncertain?
_[List questions or points that need further testing]_

---

## 8. Next Steps
Based on this experiment, I want to test next:

1. ________
2. ________
3. ________

---

## Quick Checklist
- [ ] Research question is clear
- [ ] Only changing one variable at a time
- [ ] Set up control group
- [ ] Recorded all observations
- [ ] Repeated test at least 3 times
- [ ] Documented unexpected situations
- [ ] Summarized patterns or conclusions

---

**Experiment Notes** (Free recording area):

_[Any additional thoughts, ideas, questions]_
\end{lstlisting}

%% file: appendices/app_knowledge_book.tex
\providecommand{\tightlist}{%
  \setlength{\itemsep}{0pt}\setlength{\parskip}{0pt}}

\section{Knowledge Book Evolution}
\label{sec:appendix-knowledge-book}

This appendix documents the evolution of the knowledge consolidation format used by the Scientist Agent to distill discoveries into a shared Knowledge Book. We present three formats, each with its generation prompt and an example output.

\textbf{Note on example scope.} The example outputs below are drawn from a broader Minecraft redstone exploration session that covers the full component set (including comparators, pistons, and observers). While the benchmark evaluation (Section~\ref{sec:Experiments}) restricts the component palette to basic redstone (button, wire, repeater, torch, lamp), the knowledge book format is designed to be general-purpose. These examples are included to illustrate the format's expressiveness; they are \emph{not} outputs from benchmark task runs.

\subsection{Self-determined Summary}
\label{sec:knowledge-book-self-determined}

This baseline asks the model to summarize the experiment findings without enforcing a fixed template.

\subsubsection{Generation Prompt}
\begin{lstlisting}
Summarize the discovered knowledge from these experiments.
\end{lstlisting}

\subsubsection{Example Output}
\begin{lstlisting}
Key discoveries (self-determined summary):

- Redstone dust transmits a signal strength in [0,15]. Starting from 15,
  it decays by 1 per dust block of path length; repeaters refresh the
  signal back to 15.
- Redstone dust connects only to the four cardinal neighbors (N/S/E/W);
  diagonal dust does not connect directly. Many layouts must use
  Manhattan routing.
- Repeaters introduce delay (1-4 redstone ticks) and can be chained; total
  delay is the sum of repeater delays along the path.
- In branching networks, signal strength into a T-junction is shared to
  all outgoing dust. To maximize reach, boost before a split; to control
  branch timing independently, boost after the split.
- For simultaneous activation across unequal path lengths, the longest
  path sets the minimum achievable latency; add compensating delay to
  shorter paths to synchronize edges.
- Solid blocks can be powered in ways that activate adjacent mechanisms
  but do not always propagate to adjacent dust; repeaters/comparators are
  reliable for transmitting power through blocks.
- To generate short, precise pulses from longer inputs (e.g., button
  press), normalize to an edge-triggered pulse and then extend/shaping
  with a small delay network.
\end{lstlisting}

\subsection{Finding-Explanation-Example Format}
\label{sec:knowledge-book-fee}

This format structures each discovery into a reusable finding, an explanation with constraints, and a minimal example.

\subsubsection{Generation Prompt}
\begin{lstlisting}
For each distinct discovery or significant conclusion, generate a knowledge book entry strictly following this format:

### Finding
One sentence stating the rule/behavior as a crisp claim.

### Explanation
Explain how/why the finding works and when it applies.
- Include quantitative details, constraints, edge cases, and failure modes.

### Example
Provide one minimal concrete example that demonstrates correct usage (layout, coordinates, pseudo-circuit, or scenario).

**Instructions:**
1. **Distill:** Do not just copy text. Extract the core truth from experiences.
2. **Structure:** Maintain the Finding/Explanation/Example structure for *every* entry.
3. **Coverage:** Ensure all technical details needed for reuse are captured.
4. **Clarity:** Use clear, professional technical language.
\end{lstlisting}

\subsubsection{Example Output}
\begin{lstlisting}
### Finding
Diagonal placement allows for compact star topologies.

### Explanation
Redstone dust strictly connects to the four cardinal neighbors (North, South, East, West). It does not connect diagonally.
*   **Observation:** Placing dust at `(x, z)` and `(x+1, z+1)` results in two isolated piles of dust.
*   **Correction:** Connections must be made via "Manhattan" or "Taxicab" paths (e.g., `(x, z) -> (x+1, z) -> (x+1, z+1)`).
*   **Impact:** A visual "Star" with diagonal lines is impossible. Circuits must use "Manhattan Star" shapes (cross shapes), which consume more space and can introduce path length differences if not strictly symmetrical.

---

### Finding
Signal strength decay limits branch reach; repeaters are required for long-distance transmission and must be strategically placed at T-junctions.

### Explanation
Redstone signal strength ($SS$) starts at 15 and decays by 1 per block. $SS_{next} = SS_{current} - 1$.
*   **Junction Mechanics:** A signal splitting at a T-junction provides the *same* signal strength to all branches. If a wire arrives at a junction with strength 1, it can power adjacent dust in 3 directions, but those new dust blocks will have strength 0 (signal dies).
*   **Repeater Placement:** To maximize reach after a junction, place a repeater *immediately* after the split on each branch, or *immediately* before the split on the trunk.
    *   *Before Split:* 1 Repeater powers 3 branches. Efficient, but all branches share the delay.
    *   *After Split:* Requires 3 Repeaters. Allows individual delay control per branch.

### Example
**Trunk with Pre-Junction Boost**
Maximizes signal strength entering the junction.
```text
Place components at the following coordinates:
Source: vec3(0, 64, 0)
Wire segment: vec3(0, 64, 1) to vec3(0, 64, 14)
Repeater: vec3(0, 64, 15)
Junction: vec3(0, 64, 16)
Branches: vec3(-1, 64, 16), vec3(1, 64, 16)
Lamps: vec3(-1, 64, 17), vec3(0, 64, 17), vec3(1, 64, 17)
```

**Multi-Level T-Branching**
For reaching off-axis lamps at varying distances.
```text
Layout:
vec3(10, 5, 10): Trunk Wire
vec3(11, 5, 10): Junction
vec3(11, 5, 9): Branch A Start -> vec3(11, 5, 8): L1
vec3(12, 5, 10): Repeater
vec3(13, 5, 10) to vec3(20, 5, 10): Branch B Wire -> vec3(21, 5, 10): L2
```

---

### Finding
Sequential activation is achieved by chaining repeaters with specific delay settings, effectively creating a "Timeline" of events.

### Explanation
Redstone Repeaters have configurable delays: 1, 2, 3, or 4 redstone ticks.
*   **Cumulative Delay:** Delays in series add up. $T_{total} = \sum \delta_{repeaters}$.
*   **Parallel Stages:** To trigger events at absolute times $t=0, t=2, t=6$:
    *   Lamp 1: Direct connection ($t=0$).
    *   Lamp 2: Branch -> R(2) -> Lamp ($t=2$).
    *   Lamp 3: Branch -> R(2) -> R(4) -> Lamp ($t=6$).
*   **Non-Uniform Sequences:** For patterns like [2, 1, 3], you cannot just rely on a single chain. You may need independent delay lines branching from the source if the timing isn't strictly cumulative or if geometry prevents chaining.

### Example
**Sequential Chain (Cumulative)**
Target: L1(t=0), L2(t=2), L3(t=6)
```text
Nodes:
n0: vec3(0,0,0) [Source]
n1: vec3(0,0,1) [Wire] -> vec3(1,0,1) [L1]
n2: vec3(0,0,2) [Repeater, delay=2]
n3: vec3(0,0,3) [Wire] -> vec3(1,0,3) [L2]
n4: vec3(0,0,4) [Repeater, delay=4]
n5: vec3(0,0,5) [L3]
```

**Parallel Delay Lines (Independent)**
Target: L1(t=2), L2(t=4) (Spatially separated)
```text
vec3(50, 60, 50): Source
vec3(50, 60, 51): Wire Junction
Path 1: vec3(49, 60, 51) [R(2)] -> vec3(48, 60, 51) [L1]
Path 2: vec3(51, 60, 51) [R(4)] -> vec3(52, 60, 51) [L2]
```

---

### Finding
Delay compensation synchronizes outputs across unequal path lengths by "slowing down the fast paths."

### Explanation
When paths differ in length, the longest path (requiring the most repeaters for signal strength) sets the minimum system latency ($Latency_{min}$). 
*   **The Algorithm:** 
    1.  Calculate $Delay_{inherent}$ for every path (ticks from mandatory repeaters needed for distance).
    2.  Find $Max(Delay_{inherent})$.
    3.  For every other path $i$, add compensation repeaters: $\delta_{add} = Max(Delay) - Delay_i$.
*   **Slack:** Sometimes you intentionally increase the delay of *all* paths to a higher common multiple to make the math easier (e.g., synchronize everything to 10 ticks).

### Example
**Equal-Delay Distribution Logic**
*   Path A (20 blocks): Needs 1 Repeater (min 1 tick). Delay=1.
*   Path B (5 blocks): Needs 0 Repeaters. Delay=0.
*   Path C (35 blocks): Needs 2 Repeaters (min 2 ticks). Delay=2.

*Target Delay:* 2 ticks (dictated by Path C).
*Adjustments:*
*   Path A: Add 1 tick (Total 2).
*   Path B: Add 2 ticks (Total 2).
*   Path C: Add 0 ticks (Total 2).

```text
Network Graph:
Edge(vec3(0,0,0), vec3(10,0,0)) -> Contains R(1) + R(1) -> L3
Edge(vec3(0,0,0), vec3(0,0,10)) -> Contains R(1) + R(1) -> L1
Edge(vec3(0,0,0), vec3(5,0,5))  -> Contains R(2) -> L2
```

---

### Finding
Monostable circuits and Pulse Limiters are required to generate precise pulse durations shorter than the input source.

### Explanation
A stone button generates a long pulse (~10-15 redstone ticks). To control exact duration $\tau$:
1.  **Input Normalization:** First, convert the button press into a standardized 1-tick pulse using a **Rising Edge Detector**.
2.  **Pulse Shaping:** Extend that 1-tick pulse to exactly $\tau$ ticks.
    *   *Small $\tau$ (1-4):* Use a repeater set to $\tau$ merging with the original signal? No, simpler: The 1-tick pulse powers a repeater chain that "holds" the line.
    *   *Medium $\tau$ (4-10):* Use a **Pulse Extender**. A parallel bank of repeaters is precise.
    *   *Analog Method:* A Comparator Decay Loop extends pulses but is harder to tune to exact ticks.

### Example
**Precise Pulse Generator (Target $\tau$ = 4 ticks)**

**Step 1: Monostable (Rising Edge)**
Generates a 1-tick pulse regardless of button hold time.
```text
Circuit Topology:
vec3(x,y,z) [Button] connects to:
  1. vec3(x+1,y,z) [Torch] -> vec3(x+2,y,z) [Wire]
  2. vec3(x,y,z+1) [Repeater-2] -> vec3(x+1,y,z+1) [Wire] -> vec3(x+2,y,z+1) [Connect to 1]
  Output at vec3(x+3,y,z) through Inverter.
```

**Step 2: Pulse Extension Bank (The "Timeline" method)**
To output exactly 4 ticks from a 1-tick trigger:
Input splits into 4 parallel lines of delay 1, 2, 3, 4, all merging into Output.
```text
Parallel Array:
Input: vec3(0,0,0)
Branches:
  vec3(1,0,0): R(1)
  vec3(1,0,1): R(2)
  vec3(1,0,2): R(3)
  vec3(1,0,3): R(4)
Merge: vec3(2,0,0..3) connected to Output Line vec3(3,0,0).
```

---

### Finding
Spatial management and component isolation are critical for high-density wiring.

### Explanation
As N increases, wires inevitably cross or run parallel.
*   **Interference:** Two lines of redstone dust placed side-by-side will connect.
*   **Isolation Techniques:** 
    1.  **Block Cutting:** Place a solid block between parallel wires.
    2.  **Repeater Tunneling:** Use repeaters to push signal *through* a block, allowing a perpendicular wire to run on top of that block without connecting.
    3.  **Vertical Stacking:** Run one bus line at Y=64 and another at Y=66.
*   **Slabs/Glowstone:** Use transparent blocks to run wire vertically up without cutting the signal.

### Example
**High-Density Bus Routing**
Running 3 parallel signals in a 3-wide space:
```text
Grid Configuration:
Column 0: vec3(0, y, z) -> Signal 1
Column 1: vec3(1, y, z) -> Insulator
Column 2: vec3(2, y, z) -> Signal 2
Column 3: vec3(3, y, z) -> Insulator
Column 4: vec3(4, y, z) -> Signal 3
```

**Crossing Wires (The Bridge)**
Signal A needs to cross Signal B.
```text
Layered Intersection:
Signal A Path: vec3(10, 64, 10) -> vec3(10, 65, 11) [Block Top] -> vec3(10, 64, 12)
Signal B Path: vec3(9, 64, 11) -> vec3(10, 64, 11) [Tunnel] -> vec3(11, 64, 11)
```

---

### Finding
Debug procedure for timing mismatches involves "Tick Counting" and "Edge Observation."

### Explanation
When a circuit fails simultaneity or sequence tests:
1.  **Tick Counting:** Manually trace the path from Source to Lamp, summing the delays of every repeater. Remember: Dust = 0, Torch = 1, Comparator = 1, Repeater = Configured (1-4).
2.  **Edge Observation:** Watch the *activation* (Rising Edge). Sometimes lamps turn ON simultaneously but turn OFF at different times. The contract usually specifies activation time $|t_i - t_j|$.
3.  **Ghost Power:** Ensure blocks aren't being "quasi-powered" or powered indirectly by adjacent strong-powered blocks, which can bypass intended delays.

### Example
**Debugging Table Construction**
```text
Log Data:
Target: vec3(10,10,10) [L1], Delta=6, Actual=6 (OK)
Target: vec3(20,10,10) [L2], Delta=6, Actual=7 (FAIL) -> Components at vec3(25,10,10) [R1], vec3(26,10,10) [R4], vec3(27,10,10) [R2]
```

```
\end{lstlisting}

\subsection{Claim-Proof-Constraints-Example Format}
\label{sec:knowledge-book-cpce}

The final format enforces scientific rigor by recording a law-like claim, evidence-based proof, validity constraints, and a concrete application example.

\subsubsection{Generation Prompt}
\begin{lstlisting}
For each distinct discovery or significant conclusion, generate an entry in the knowledge book strictly following this structure:

### 1. Claim (Law)
State the finding as a universal, law-like statement (similar to a mathematical theorem or physical law). It must be concise, precise, and assertive.
*   *Bad:* "Redstone gets weaker as it goes far."
*   *Good:* "The Law of Linear Signal Decay: Redstone signal strength $S$ decreases by exactly 1 unit per block of distance $d$, such that $S_{d} = S_{initial} - d$."

### 2. Proof
Provide the rigorous logic that supports this claim. This acts as the "proof."
*   Synthesize specific evidence from the experiments.
*   Connect prior established findings with new data.
*   Outline the deductive steps that lead inevitably to the claim.
*   Maintain a high standard of rigor; mere observation is not proof--explain *why* the observation confirms the law.

### 3. Constraints
Identify potential weaknesses or exceptions in the proof.
*   Are there edge cases where the law fails? (e.g., "Does not apply across chunk boundaries in specific versions.")
*   Are there assumptions made that require further validation?
*   Note any observed anomalies that contradict the main claim.

### 4. Example
Demonstrate the utility of this finding.
*   Provide a concrete example or usage scenario.
*   Describe a mechanism or design pattern that relies specifically on this law being true.
\end{lstlisting}

\subsubsection{Example Output}
\begin{lstlisting}
### 1. The Law of Linear Signal Attenuation

#### Claim
Redstone signal strength $S$ decays linearly by exactly 1 unit for every block of distance traversed $d$, governed by the equation $S_{d} = S_{source} - d$. The maximum transmission range without amplification is 15 blocks, occurring when $S_{source} = 15$.

#### Proof
*   **Observation:** In all experiments, placing a redstone wire adjacent to a power source (Strength 15) results in a strength of 14 on the wire.
*   **Deduction:** The signal terminates (Strength 0) exactly at the 16th block.
*   **Geometry:** Distance is calculated using "Taxicab" or "Manhattan" geometry ($\Delta x + \Delta z$), not Euclidean distance. A wire spiraling or changing elevation consumes signal strength identical to a straight line of the same block count.

#### Constraints
*   **Constraint:** This law applies strictly to Redstone Dust. Solid blocks powered by repeaters or comparators do not degrade the signal *within* the component itself, but the dust exiting them restarts the decay.
*   **Edge Case:** Signal strength does not decay when passing through a Comparator in comparison mode (it maintains the input strength).

#### Example
*   **Trunk with Pre-Junction Boost:** Place a repeater *immediately* before a trunk line splits into multiple branches. This ensures all branches receive maximum signal strength (15) and resets the decay counter for the new paths.
    ```text
    B (SS=15)
    |
    [14 blocks of wire]
    |
    R (Repeater, Boosts to SS=15)
    |
    +-------+-------+
    |       |       |
    W(15)   W(15)   W(15) <-- All branches start fresh
    |       |       |
    L1      L2      L3
    ```
*   **Multi-Level T-Branching:** When branching to off-axis lamps at varying distances, independent repeaters are required *after* the split to ensure the longer branch receives sufficient power without affecting the timing of the shorter branch.
    ```text
    Trunk --+-- Branch A (Short) -- L1
            |
            R (Repeater needed to reach Branch B)
            |
            +-- Branch B (Long) -- L2
    ```
*   **Signal Strength Reader:** A compact circuit using comparators to output a different signal strength based on the input signal, allowing for more granular information transfer than simple on/off.
    ```text
    Input (SS=0-15) --> Comparator (Subtract Mode) --> Output (SS varies based on input)
    ```

---

### 2. The Principle of Quasi-Connectivity (Action at a Distance)

#### Claim
Certain activation components (Pistons, Dispensers, Droppers) behave as if they are two blocks high for the purpose of receiving power. A component at $(x, y, z)$ will activate if a signal would validly power a theoretical component at $(x, y+1, z)$, even if the space at $y+1$ is empty or air.

#### Proof
*   **Experiment:** A piston placed at $y$ activates when a redstone block is placed at $y+2$, provided there is no solid block blocking the "sight" (though air transmits it).
*   **Logic:** The game code checks for power in the block *above* the component to support door mechanics, but this check was generalized, resulting in "BUD" (Block Update Detector) behavior where the component receives power but doesn't realize it until a neighbor updates.

#### Constraints
*   **Constraint:** This is strictly a **Java Edition** mechanic. It does not function in Bedrock Edition.
*   **Anomaly:** The component often requires a "block update" (neighbor changing state) to realize it has been quasi-powered or quasi-depowered. This leads to "budding" where a piston stays extended even after power is cut.

#### Example
*   **Quasi-Connectivity Visualization:** Powering the space *above* a piston activates it, even with an air gap.
    ```text
    [Redstone Block]  <-- Power Source
          |
        (Air)         <-- "Invisible" connection range
          |
    [Piston/Dropper]  <-- Activates via QC
    ```
*   **Compact Piston Walls:** Powering a wall of pistons from the ceiling to avoid messy floor wiring.
    ```text
    [Redstone Dust Line]
    [Solid Block]
    [Piston] [Piston] [Piston] <-- All activate from blocks above
    ```
*   **BUD Switch (Block Update Detector):** A piston or dropper configured to be quasi-powered. It only extends/activates when a block *adjacent* to it is updated, even if the power source itself hasn't changed. This is used for detecting block changes or growth.
    ```text
          (QC Power Source - e.g., Redstone Block 2 blocks above)
          |
    [Piston]  <-- Quasi-powered, but not updated
    [Any Block] <-- When this block updates (e.g., placed/broken), Piston extends
    ```

---

### 3. The Theorem of Orthogonal Transmission (Manhattan Connectivity)

#### Claim
Redstone dust connects exclusively to the four cardinal neighbors (North, South, East, West). Diagonal transmission is impossible directly; all diagonal signal propagation must approximate the hypotenuse via a stepped "Manhattan" path, where $Distance_{steps} = |\Delta x| + |\Delta z|$.

#### Proof
*   **Observation:** Placing dust at $(x, z)$ and $(x+1, z+1)$ results in two isolated components.
*   **Logic:** To connect them, dust must be placed at either $(x+1, z)$ or $(x, z+1)$.
*   **Consequence:** A visual "star" topology is topologically impossible. All branching must occur at orthogonal T-junctions.

#### Constraints
*   **Exception:** Vertical transmission via "soft" transparent blocks (slabs, glowstone) allows a signal to travel diagonally *upward* in a zigzag pattern without connecting to adjacent horizontal wires, effectively creating a diode.

#### Example
*   **High-Density Bus Routing:** Parallel wires must be separated by blocks (e.g., `Wire-Block-Wire`) to prevent interference.
    ```text
    Top View:
    W1  B   W2  B   W3
    |   |   |   |   |
    W1  B   W2  B   W3
    ```
*   **Crossing Wires (The Bridge):** To cross Signal A over Signal B, A must be routed over a block while B passes underneath. This "Bridge" uses the vertical orthogonality to separate planar signals.
    ```text
          A
          |
          W
          |
    B -- [Block] -- B  <-- Signal A goes over Signal B
          |
          W (Signal B runs under the block)
          |
          A
    ```
*   **Compact Corner Turns:** Instead of a wide curve, redstone dust can turn 90 degrees using a single corner block, adhering to the orthogonal connection rule. This is fundamental for compact layouts.
    ```text
    W -- W
         |
         W
    ```

---

### 4. The Law of Solid Block Conductivity (Hard vs. Soft Power)

#### Claim
A solid (opaque) block receiving a redstone signal becomes "weakly powered," capable of activating adjacent mechanism components (lamps, pistons) but *not* adjacent redstone dust. To power adjacent dust, the block must be "strongly powered" by a repeater, comparator, or torch.

#### Proof
*   **Experiment:** Wire pointing into a Stone block activates a lamp touching the Stone, but does not activate a wire continuing from the other side of the Stone.
*   **Correction:** Replacing the input wire with a Repeater pointing into the Stone activates the output wire.

#### Constraints
*   **Transparency Exception:** Transparent blocks (Glass, Slabs, Glowstone) *cannot* be powered at all. They are insulators. Wire placed on them transmits signal, but the block itself does not hold a charge.

#### Example
*   **Strong vs. Weak Power:** Differentiating how signal propagates through blocks.
    ```text
    (Weak Power - Fails to continue)
    Wire --> [Block] -/-> Wire (OFF)
    
    (Strong Power - Successfully transmits)
    Repeater --> [Block] ---> Wire (ON)
    ```
*   **Repeater Tunneling:** Using a repeater to push signal *into* a block allows a perpendicular wire to run *on top* of that block without connecting to the signal below.
    ```text
          W (Wire on top)
          |
    R --> [Block] --> R (Signal passes THROUGH block)
          |
          W (Wire on top)
    ```
*   **Redstone Lamp Array:** Activating multiple redstone lamps in a compact space by strongly powering a row of solid blocks, each adjacent to a lamp.
    ```text
    R (Repeater) --> [Block] --> [Block] --> [Block]
                     |           |           |
                    [Lamp]      [Lamp]      [Lamp]
    ```

---

### 5. The Principle of Instantaneous Translocation (Instant Wire)

#### Claim
The movement of a block by a piston is processed as an instantaneous event ($t=0$ ticks) regarding the connectivity of attached components. This allows for "Instant Wire" where a signal travels infinite distance in the same game tick.

#### Proof
*   **Mechanism:** When a piston extends, it pushes a block. The game engine updates the new position of the block and its neighbors *within the same tick* logic cycle.
*   **Setup:** A chain of pistons, where Piston A pushing Block A completes a circuit for Piston B, will ripple through the entire chain before the next tick is processed.

#### Constraints
*   **Constraint:** Requires precise sub-tick update order (update suppression or specific block placement order) to function reliably in all directions.
*   **Cost:** High noise and potential for client-side visual desync (ghost blocks).

#### Example
*   **Piston Instant Wire Chain:** A mechanical cascade where moving blocks complete circuits instantly.
    ```text
    In -> [Piston] -> [Block A]
                         |
       (Block A moves here to complete circuit)
                         |
                   [Redstone Dust] -> [Piston] -> [Block B] -> Out
    ```
*   **Long-Distance Interrupt:** Transmitting a "Stop" signal to a farm 500 blocks away instantly.
*   **T-Flip Flop (Instant Set/Reset):** Using two sticky pistons pushing a redstone block back and forth to create a compact T-Flip Flop where a single pulse instantly toggles the output state, leveraging instantaneous block updates.
    ```text
    In -> Piston (Sticky) -> [Redstone Block] <- Piston (Sticky) <- In
                                     |
                                   Output
    ```

---

### 6. The Law of Discrete Quantized Time (Tick Logic)

#### Claim
Time in Redstone circuits is quantized into discrete "Redstone Ticks" (equivalent to 0.1 seconds). Events cannot occur "between" ticks. Total latency in a serial circuit is the arithmetic sum of all individual component delays: $T_{total} = \sum \delta_{components}$.

#### Proof
*   **Mechanics:** Repeaters have configurable delays of 1, 2, 3, or 4 ticks.
*   **Observation:** A chain of two repeaters set to 2 and 4 results in an exact 6-tick delay.
*   **Logic:** Parallel paths with unequal tick sums arrive at different times. The "slowest" path determines the minimum latency for a synchronized system.

#### Constraints
*   **Sub-Tick Events:** As noted in Law 5, piston events can resolve within a single tick, effectively having $\delta=0$.
*   **Pulse Chopping:** If an input pulse is shorter than a repeater's delay setting (e.g., 1-tick pulse into 4-tick repeater), the repeater effectively "extends" or stabilizes the pulse to its minimum duration in some versions, or ignores it in others.

#### Example
*   **Sequential Activation:** Creating "Waves" of activation (e.g., runway lights) by chaining repeaters to trigger lamps at $t, t+2, t+4$.
    ```text
    B --+-- L1 (t=0)
        |
        R(2)
        |
        +-- L2 (t=2)
        |
        R(4)
        |
        L3 (t=6)
    ```
*   **Parallel Delay Lines:** Independent delay lines can be used when spatial separation prevents a single chain.
    ```text
          +-- R(2) -- L1
          |
    B -- -+
          |
          +-- R(4) -- L2
    ```
*   **Equal-Delay Distribution Logic:** Synchronizing a "Fast" path with a "Slow" path by adding filler repeaters.
    ```text
    Path C: B -- R(1) -- W -- R(1) -- W -- L3 (Native 2 ticks)
    Path A: B -- R(1) -- W -- R(1) -- W -- L1 (1 tick for distance + 1 tick compensation)
    Path B: B -- R(2) -- W -- L2              (2 ticks purely for compensation)
    ```
*   **Pulse Shaping:** Using Monostable circuits (Rising Edge Detectors) to convert variable-length button presses into precise 1-tick pulses.
    ```text
          +-- T (Torch) --+
          |               |
    B -- -+-- R(2) -------+-- NOT -- (Output)
    ```
\end{lstlisting}

%% file: appendices/app_failure_taxonomy.tex
\section{Qualitative Failure Analysis}
\label{app:failure_taxonomy}

To ground our quantitative gap decomposition (Section~\ref{sec:experiment_design}) in concrete agent behavior, we constructed 12 device variants of the 32-lamp simultaneous activation task (Family~A), each exhibiting a distinct failure mode.
A working reference device (Case~W, 32/32 lamps lit) serves as the baseline.
The 12 failure cases are organized into three categories based on the \emph{stage} of the signal pipeline that breaks, progressing from the power source to the lamps.

\subsection{Representative Failure Cases}

\begin{figure*}[h]
  \centering

  \begin{subfigure}[t]{0.48\linewidth}
    \centering
    \includegraphics[width=\linewidth]{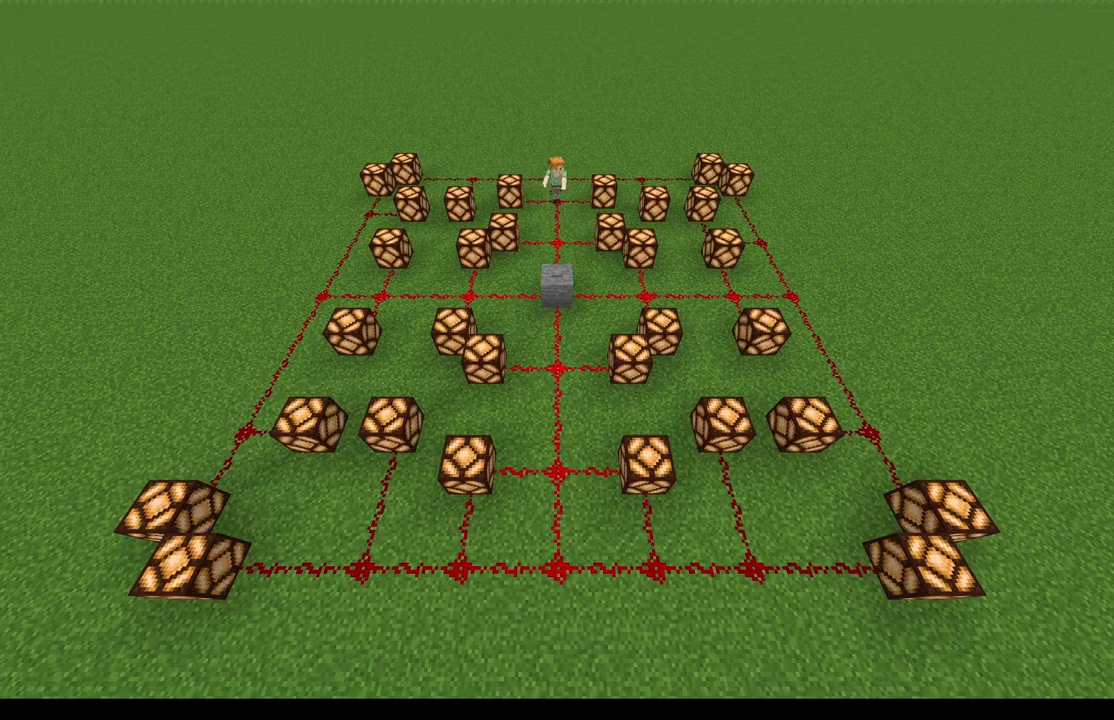}
    \caption{\textbf{Working Reference} (32/32 lamps lit). Correct four-axis topology with branch wires delivering signal to all lamps.}
  \end{subfigure}\hfill
  \begin{subfigure}[t]{0.48\linewidth}
    \centering
    \includegraphics[width=\linewidth]{figures/assets/case-9-backwards-repeaters_lit.jpg}
    \caption{\textbf{Backwards Repeaters} (8/32 lamps lit). Four repeaters placed facing inward block outgoing signal beyond distance 4.}
  \end{subfigure}

  \vspace{0.8em}

  \begin{subfigure}[t]{0.48\linewidth}
    \centering
    \includegraphics[width=\linewidth]{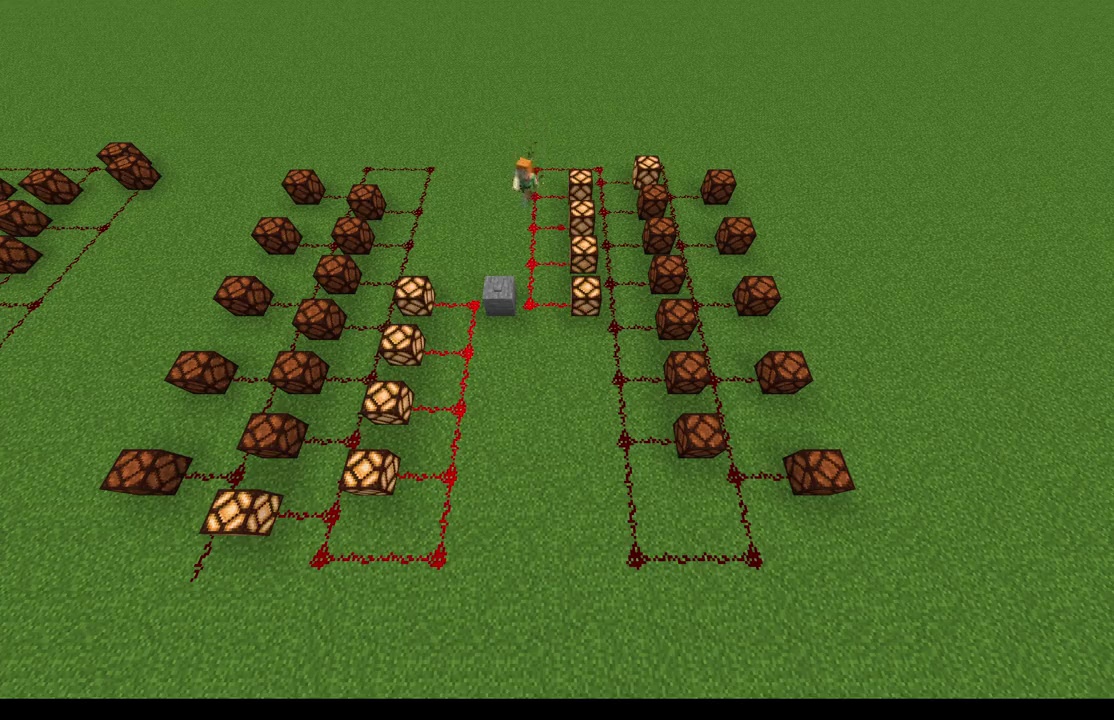}
    \caption{\textbf{Long Snake} (10/32 lamps lit). A 94-wire zigzag path without repeaters; signal decays to zero after 15 blocks.}
  \end{subfigure}\hfill
  \begin{subfigure}[t]{0.48\linewidth}
    \centering
    \includegraphics[width=\linewidth]{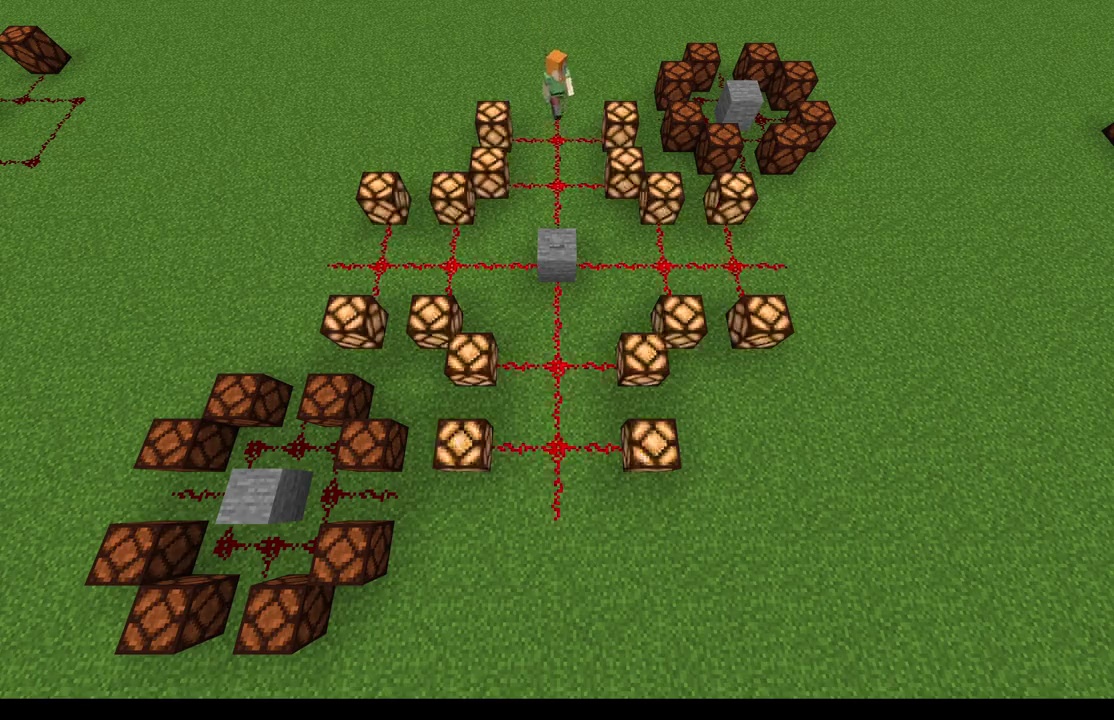}
    \caption{\textbf{Islands} (16/32 lamps lit). Two disconnected sub-circuits with no wire connection to the button. Center-connected lamps light; island lamps stay dark.}
  \end{subfigure}

  \vspace{0.8em}

  \begin{subfigure}[t]{0.48\linewidth}
    \centering
    \includegraphics[width=\linewidth]{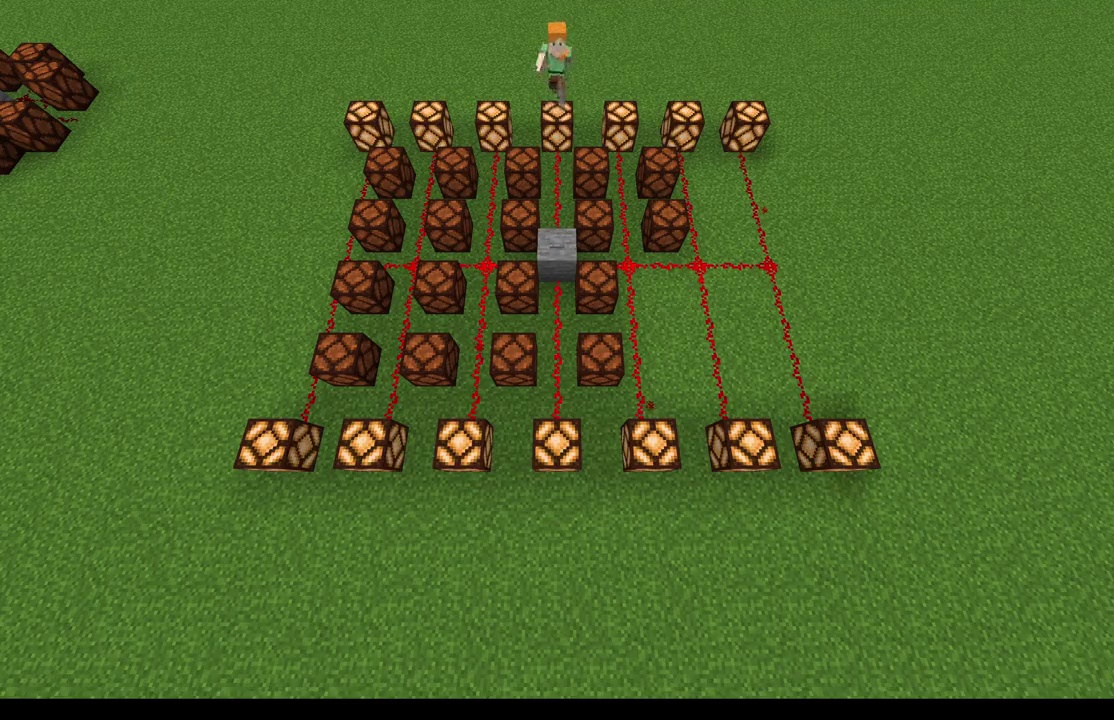}
    \caption{\textbf{Parallel Lines} (14/32 lamps lit). N--S wires carry full power but only connect north and south; lamps placed between lines stay dark due to wire directionality.}
  \end{subfigure}

  \caption{\textbf{Representative failure cases from the 32-lamp broadcast task.}
  (a)~Working device where all lamps activate simultaneously.
  (b)~\emph{Structural} failure: repeaters oriented backwards create one-way barriers.
  (c)~\emph{Signal propagation} failure: long serial path without amplification causes signal decay.
  (d)~\emph{Connectivity} failure: isolated sub-circuits receive no power from the button.
  (e)~\emph{Wire semantics} failure: directional wire connections prevent lateral power delivery.}
  \label{fig:failure_cases}
\end{figure*}

Figure~\ref{fig:failure_cases} shows five representative cases: a working reference, a structural failure (backwards repeaters), a signal propagation failure (long snake), a connectivity failure (islands), and a wire-semantics failure (parallel lines). Lit lamps appear bright; dark lamps remain unlit. These cases illustrate how agents can construct circuits that are \emph{topologically plausible} yet \emph{functionally broken}---failures that are difficult to diagnose without understanding the underlying redstone mechanics.

\subsection{Failure Taxonomy}

These failures naturally cluster into three categories that align with our capacity decomposition:

\paragraph{Category 1: Structural failures.}
These failures arise when the agent places blocks at incorrect positions, uses wrong orientations, or omits critical connections---resulting in a circuit whose physical structure is itself broken.
This category maps primarily to \textbf{knowledge application} capacity: the agent may understand redstone principles but fails to translate them into correct block placements.

\begin{itemize}[leftmargin=1.5em]
\item \textbf{Case 5 --- Broken Bridge} (0/32 lit). The 4 wires immediately adjacent to the center stone are missing, completely disconnecting the power source from the wire network. A single-point structural omission disables the entire circuit.
\item \textbf{Case 9 --- Backwards Repeaters} (8/32 lit). Four repeaters are placed with their input sides facing \emph{away} from the signal source. Repeaters are one-way devices; reversing their orientation creates barriers that block signal propagation beyond distance~4 on each axis.
\item \textbf{Case 4 --- Glass Pedestal} (0/32 lit). The center stone block is replaced with glass. Transparent blocks cannot conduct redstone power, so the button press never enters the wire network despite all 112 wires and 32 lamps being correctly placed.
\item \textbf{Case 8 --- Axes Only} (0/32 lit). Only the 4 main axis wires are placed (32 wires total) with no branch wires. Signal flows along all axes but no lamp is adjacent to any axis wire---the ``last mile'' delivery is entirely missing.
\end{itemize}

\paragraph{Category 2: Signal propagation failures.}
These failures occur in structurally connected circuits where the signal \emph{cannot reach} all target lamps due to insufficient wire coverage or missing amplification.
This category maps to \textbf{knowledge discovery} capacity: the agent has not discovered (or fails to apply) the 15-block signal decay rule.

\begin{itemize}[leftmargin=1.5em]
\item \textbf{Case 1 --- Long Snake} (10/32 lit). A 94-wire zigzag path with no repeaters. Signal starts at power~15 from the center stone and loses 1 power per wire, dying after 15 blocks. Only the nearest 10 lamps are within range.
\item \textbf{Case 6 --- Missing Rails} (24/32 lit). All perimeter distribution wires are removed (26 wires). Inner lamps are reachable via branch wires, but the 8 corner lamps lose their signal path.
\item \textbf{Case 7 --- Tiny Core} (8/32 lit). Only 20 wires (vs.\ 112 in the working device) extend 3 blocks in each direction. The 8 nearest lamps light; the remaining 24 are beyond wire reach.
\end{itemize}

\paragraph{Category 3: Wire semantics failures.}
These failures involve circuits that are structurally complete and carry sufficient signal power, yet fail because the agent does not understand redstone wire's \emph{directional connection semantics}---wire only powers blocks in the directions it visually connects to.
This category maps to \textbf{knowledge gap identification} capacity: the agent does not even recognize that wire directionality is a relevant factor, making it the most subtle and difficult failure class.

\begin{itemize}[leftmargin=1.5em]
\item \textbf{Case 3 --- Parallel Lines} (14/32 lit). Seven parallel N--S wires connected by an E--W trunk. The N--S wires carry full power but only connect north and south, so lamps placed \emph{between} lines (requiring E--W power) stay dark.
\item \textbf{Case 10 --- Connection Trap} (20/32 lit). Twelve extra ``trap'' wires are added perpendicular to existing dead-end branches, causing auto-connection to form L-shapes that redirect flow \emph{away} from lamps. Adding wires \emph{breaks} a working circuit.
\item \textbf{Case 12 --- The Ring} (0/16 lit). A closed ring of wire surrounds the center. Every ring wire connects to its two neighbors along the ring---never outward. Despite carrying power~8--15 throughout, zero power is delivered to external lamps.
\item \textbf{Case 2 --- Islands} (16/32 lit). The circuit contains three stone anchors, but only one has a button. Two ``island'' clusters are completely disconnected from the power source. The agent built plausible-looking sub-circuits that have no electrical connection to the button.
\item \textbf{Case 11 --- Dead-End Hooks} (12/20 lit). Hook wires at arm tips change the last wire's connection direction, cutting power to lamps that would otherwise be lit. A single extra wire at a dead end reverses which direction receives power.
\end{itemize}

\subsection{Summary}

Table~\ref{tab:failure_summary} summarizes all cases.
The failure taxonomy reveals a progression from obvious structural errors (Category~1) to subtle semantic misunderstandings (Category~3).
Notably, Category~3 failures produce circuits that \emph{appear correct} upon visual inspection---wires are connected, power flows through the network, yet lamps remain dark.
These cases demonstrate that the hardest failures to diagnose are those where the agent's mental model of the domain is \emph{qualitatively incomplete} rather than quantitatively inaccurate.
This progression mirrors the capacity gap hierarchy observed in our quantitative results: knowledge application failures are the most straightforward to diagnose, while identification failures require the deepest domain understanding.

\begin{table}[h]
\centering
\small
\caption{Complete failure taxonomy for the 32-lamp broadcast task.}
\label{tab:failure_summary}
\begin{tabular}{clllr}
\toprule
\textbf{Cat.} & \textbf{Case} & \textbf{Failure Mode} & \textbf{Capacity Gap} & \textbf{Lit} \\
\midrule
--- & W: Working Ref. & None (correct) & --- & 32/32 \\
\midrule
\multirow{4}{*}{1} & 4: Glass Pedestal & Transparent center block & Application & 0/32 \\
& 5: Broken Bridge & Missing center junction & Application & 0/32 \\
& 8: Axes Only & No branch wires & Application & 0/32 \\
& 9: Backwards Rep. & Reversed repeater dir. & Application & 8/32 \\
\midrule
\multirow{3}{*}{2} & 1: Long Snake & Signal decay (no amp.) & Discovery & 10/32 \\
& 6: Missing Rails & Incomplete coverage & Discovery & 24/32 \\
& 7: Tiny Core & Insufficient wire reach & Discovery & 8/32 \\
\midrule
\multirow{5}{*}{3} & 2: Islands & Disconnected sections & Identification & 16/32 \\
& 3: Parallel Lines & Wire direction mismatch & Identification & 14/32 \\
& 10: Connection Trap & Auto-connect redirect & Identification & 20/32 \\
& 11: Dead-End Hooks & Dead-end flow reversal & Identification & 12/20 \\
& 12: The Ring & Loop flow trap & Identification & 0/16 \\
\bottomrule
\end{tabular}
\end{table}